# Breath as a biomarker: A survey of contact and contactless applications and approaches in respiratory monitoring

Almustapha A. Wakili[*], Babajide J. Asaju, Woosub Jung

*CIS Department, Towson University, USA*

---

ABSTRACT

Breath analysis has emerged as a critical tool in health monitoring, offering insights into respiratory function, disease detection, and continuous health assessment. While traditional contact- based methods are reliable, they often pose challenges in comfort and practicality, particularly for long-term monitoring. This survey comprehensively examines contact-based and contactless approaches, emphasizing recent advances in machine learning and deep learning techniques applied to breath analysis. Contactless methods, including Wi-Fi Channel State Information and acoustic sensing, are analyzed for their ability to provide accurate, noninvasive respiratory monitoring.

We explore a broad range of applications, from single-user respiratory rate detection to multi- user scenarios, user identification, and respiratory disease detection. Furthermore, this survey details essential data preprocessing, feature extraction, and classification techniques, offering comparative insights into machine learning/deep learning models suited to each approach. Key challenges like dataset scarcity, multi-user interference, and data privacy are also discussed, along with emerging trends like Explainable AI, federated learning, transfer learning, and hybrid modeling. By synthesizing current methodologies and identifying open research directions, this survey offers a comprehensive framework to guide future innovations in breath analysis, bridging advanced technological capabilities with practical healthcare applications.

## 1. Introduction

Respiration is a primary physiological activity and a widely used measure of general health. Chronic respiratory diseases affect the lives of millions of people around the globe. For example, Chronic Obstructive Pulmonary Disease (COPD) affects the lungs of more than 250 million people worldwide—according to the World Health Organization (World Health Organization, 2023); asthma affects approximately 262 million people (Global Asthma Network, 2022). The early stages of these diseases can be detected or monitored with the help of pulmonology. Breath analysis has thus become a promising diagnostic tool, providing rich information on several health conditions and disease states (Xu, Guo, & Chen, 2020). Characterized by non-invasive, real-time, and possibly cost-effective features, the analysis uses VOC detection in exhaled breath. These VOCs act as biomarkers for many health issues; thus, breath analysis becomes one of the crucial aspects in the early diagnosis and continuous monitoring of diseases (Ge et al., 2022). Therefore, understanding and analyzing VOCs in exhaled breath is required to enhance overall health.

Traditionally, direct contact-based breath analysis involves the sampling of exhaled air, usually assisted by a mouthpiece connected to an analytic device (Lawal et al., 2017). The data obtained from this source is of high value and has triggered significant strides in healthcare diagnostics. However, this contact-based method involves physical contact with the patient, which can cause discomfort and increase the risk of cross-contamination. Despite these disadvantages, this remains primarily used in the clinical setting where high-tech sensors and analytical methods are applied for detecting and analyzing disease-associated biomarkers (Ge et al., 2022). However, the inherent drawbacks of this contact-based approach have motivated the development of contactless breath analysis. This new approach enables a non-invasive assessment, enhancing patients' safety and comfort by assessing respiration function without direct touch (Guan et al., 2022). This innovative method offers a non-invasive alternative, improving patient comfort and safety by monitoring respiratory activity without direct contact (Guan et al., 2022). This has transformed traditional methods by leveraging advanced technologies such as Wi-Fi, radio frequency (RF), and even machine learning, becoming a game-changer in the field. Numerous studies support This shift by demonstrating the advantages of contactless methods in various applications (Guan et al., 2022; Ma et al., 2019; Nicolo et al., 2020`; Wan et al., 2023). Further, various comprehensive reviews (Ali, Elsayed, et al., 2021; Bawua et al., 2021; Ge et al., 2022; Zarandah et al., 2023) have documented these advancements and highlighted their effectiveness, emphasizing the significant potential of Wi-Fi and RF technologies in enhancing patient comfort and monitoring accuracy.

---

* Corresponding author.
  *E-mail address:* awakili@towson.edu (A.A. Wakili).

Although many advanced technologies are used in this field, the breath analysis domain still faces a significant challenge. There is a noticeable lack of standardized application scenarios. Applications like Breath Rate Monitoring, Breath Pattern Analysis, COPD Detection, and Breath Type Classification all strive to detect human respiratory signals accurately (Zeng et al., 2018). However, despite their shared goals, these applications often work in silos, without a unified framework for methodologies and techniques (Ginsburg et al., 2018; Liebetruth et al., 2024). This fragmentation in the field underscores the urgent need for a comprehensive review to bridge these gaps and bring together insights from various studies, as emphasized by several existing surveys (Ge et al., 2022; Zarandah et al., 2023).

Table 1 highlights the contributions of our survey compared to other survey papers in this domain. For example, Ge et al. (Ge et al., 2022) review Wi-Fi sensing technology for non-invasive health monitoring, identifying challenges like privacy concerns and signal accuracy. However, their review lacks a comprehensive framework for standardizing application scenarios across different breath analysis applications. Bawua et al. (Bawua et al., 2021) provide valuable insights into visual, thoracic impedance, and electrocardiographic methods for measuring respiratory rates in hospitalized patients. However, their review does not cover the recent contactless methods such as Wi-Fi sensing and machine learning. Nicolo et al. (Nicolo et al., 2020`) investigated new sensor technologies and their applications in healthcare, highlighting the need for more accurate, reliable, and scalable sensors. Yet, they do not delve deeply into the specific challenges and solutions for breath analysis. Ginsburg et al. (Ginsburg et al., 2018) evaluate tools for measuring respiratory rates to identify childhood pneumonia, underscoring the need for reliable and easy-to-use tools in resource-limited settings. Lastly, Zarandah et al. (Zarandah et al., 2023) highlight machine and deep learning methods for detecting and classifying respiratory system diseases, pointing out the shortage of tools and publicly available datasets, which hinders the translation of academic research into industrial applications. In summary, none of the review papers offers a comprehensive overview of respiratory monitoring applications, approaches, and architectures, while our work covers a broader range of the current literature. Specifically, our survey addresses these gaps by analyzing the current literature and aiming to achieve the following objectives.

**Table 1**
**Related surveys and review papers Vs. Our survey paper**.

| Ref | Key Topic | Recent Approaches Covered | | | | Breath Applications Discussed | | | | | Common Architecture Discussed | | |
|---|---|---|---|---|---|---|---|---|---|---|---|---|---|
| | | W-D[a] | W-CSI[b] | A-S[c] | R-T[d] | SU-RR[e] | MU-RR[f] | UID[g] | HR[h] | OH[i] | D-Ex[j] | Pre-P[k] | Algo[l] |
| Ge et al. (2022) | Wi-Fi Sensing for Future Healthcare | ✗ | ✓ | ✗ | ✗ | ✓ | ✓ | ✓ | ✓ | ✓ | ✗ | ✓ | ✓ |
| Ali, Elsayed, et al. (2021) | Review of Contact and Contactless Breathing Rate Monitoring Systems | ✓ | ✗ | ✓ | ✓ | ✓ | ✓ | ✗ | ✗ | ✓ | ✓ | ✓ | ✓ |
| Ginsburg et al. (2018) | Respiratory Rate Tools for Pneumonia Detection | ✓ | ✗ | ✗ | ✗ | ✓ | ✗ | ✗ | ✗ | ✓ | ✗ | ✓ | ✓ |
| Bawua et al. (2021) | Electro-cardiographic Methods for Respiratory Rate Measurement | ✓ | ✗ | ✗ | ✗ | ✓ | ✗ | ✗ | ✗ | ✓ | ✗ | ✗ | ✗ |
| Zarandah et al. (2023) | Machine and Deep Learning for Respiratory Diseases Detection | ✓ | ✗ | ✓ | ✗ | ✗ | ✗ | ✗ | ✗ | ✓ | ✓ | ✓ | ✓ |
| Ma et al. (2019) | Wi-Fi Sensing for Vital Information | ✗ | ✓ | ✗ | ✗ | ✓ | ✓ | ✗ | ✓ | ✗ | ✓ | ✓ | ✓ |
| Nicolo et al. ` (2020) | Importance of RR Monitoring from Healthcare to Sport and Exercise | ✓ | ✗ | ✗ | ✗ | ✗ | ✗ | ✓ | ✓ | ✗ | ✗ | ✗ | ✓ |
| Liebetruth et al. (2024) | Review of HR and RR Measurements using Radar Tech | ✗ | ✗ | ✗ | ✓ | ✓ | ✓ | ✗ | ✗ | ✓ | ✗ | ✓ | ✓ |
| This Paper | Contact and Contactless Breath Analysis | ✓ | ✓ | ✓ | ✗ | ✓ | ✓ | ✓ | ✓ | ✓ | ✓ | ✓ | ✓ |

| | |
|---|---|
| a | Wearable Devices Approaches. |
| b | Wi-Fi CSI Approach. |
| c | [c] Acoustic Signal Approach. |
| d | [d] Radar Technology Approach. |
| e | [e] Single User Respiratory Rate Detection Applications. |
| f | [f] Multiple Users Respiratory Rate Detection Applications. |
| g | User Identification Applications. |
| h | [h] Heart Rate Detection and Monitoring Applications. |
| I | Other Healthcare Monitoring and Disease Detection Applications. |
| j | Data Extraction Technique. |
| k | Data Pre-Processing. |
| l | Algorithms for Data Processing. |

(1) Standardizing Application Scenarios: We review various approaches used in breath analysis applications, providing a comprehensive framework for sharing methodologies and techniques. This addresses the need for standardized application scenarios identified by (Ge et al., 2022) and (Liebetruth et al., 2024).

(2) Exploring Common and Specific Techniques: We offer a detailed comparison by examining standard application methodologies and techniques. This extends the analysis provided by (Ali, Elsayed, et al., 2021) and (Bawua et al., 2021), which focus on comparative analysis but do not explore hybrid methods that combine contact-based and remote technologies.

(3) Advanced Technologies: We include newer, contactless methods, such as Wi-Fi sensing and machine learning, which have not been thoroughly explored in prior reviews. This builds on the works (Nicolo et al., 2020˙ ) and (Zarandah et al., 2023). Providing a more comprehensive analysis of their applications in healthcare settings.

(4) Providing a Focused Analysis on Key Challenges and Emerging Trends: Our survey delves into the key challenges and emerging trends in breath analysis technologies, filling the gaps highlighted by (Ma et al., 2019) (Ginsburg et al., 2018).

(5) Identifying Open Research Problems and Untapped Areas: We build on the findings from (Liebetruth et al., 2024) by identifying open research problems and untapped areas, particularly the applications of AI in breath analysis.

(6) Expanding the Scope to Non-clinical Settings: Our survey expands the scope to include nonclinical settings and various breath analysis applications, offering a more comprehensive field view. This complements the clinical focus of (Ginsburg et al., 2018).

As a complementary overview, Fig. 1 provides an overview of the main applications and approaches in breath analysis covered by this survey. The diagram highlights the primary categories of applications-single-user and multi-user respiratory rate detection, user identification, heartbeat monitoring, and disease detection—and their corresponding methods. These methods are classified into contact-based and contactless approaches, such as wearable devices, Wi-Fi CSI, and acoustic signal processing.

This survey will cover several sections that address various facets of breath analysis. In Section 2, the backdrop of breath analysis, including its historical context and technological developments—will be discussed. Section 3 will focus on breath analysis applications and their significant contributions. Section 4 will examine the techniques and methodologies used in breath analysis. Section 5 will cover the common architecture of breath analysis systems in detail. Section 6 will explore obstacles, constraints, and future directions. The survey will conclude in Section 7, summarizing the main conclusions and providing suggestions for further research.

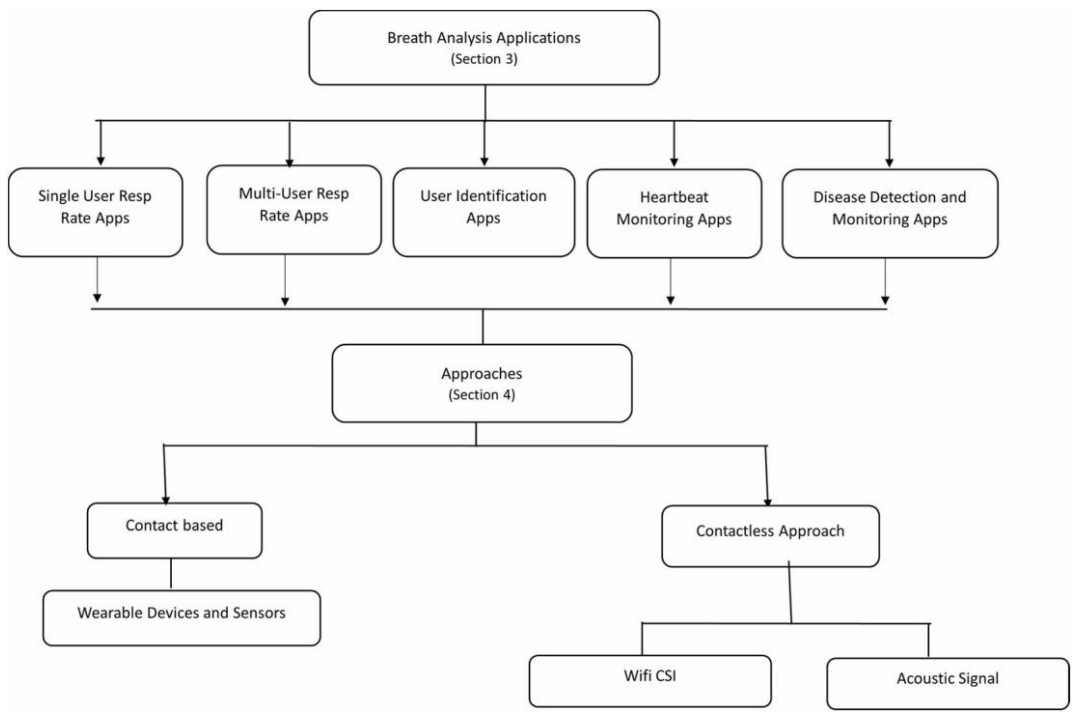

**Fig. 1.** Mind map for Breath Analysis Applications and Approaches covered by this survey.

## 2. Background

### 2.1. Evolving trends in respiratory monitoring

Breath analysis has long been recognized as a fundamental tool in healthcare. By detecting respiratory patterns, breath analysis offers insights into various health conditions. It allows healthcare professionals to assess respiratory health and detect early signs of chronic obstructive pulmonary disease (COPD), asthma, and metabolic disorders. Traditionally, breath analysis methods have been contact-based, relying on direct patient interaction for accurate data collection (Ates & Dincer, 2023; Chen et al., 2021).

Traditional methods of breath analysis using contact-based techniques typically require wearable sensors placed on the chest or abdomen. These sensors detect physiological signals related to breathing, such as chest movements and respiratory rates. For instance, Imtiaz et al. (Imtiaz, 2021) employ photoplethysmography (PPG) sensors to monitor respiratory activity continuously. Additionally, stethoscopes remain the standard tool in clinical settings for analyzing lung sounds (Seah et al., 2023). While effective, these methods tend to be invasive, costly, and less suitable for long-term monitoring, prompting the development of alternative, non-invasive diagnostic methods.

A significant shift in respiratory monitoring technologies is the emergence of contactless breath analysis. By leveraging wireless technologies such as Wi-Fi, radio frequency (RF), and radar, contactless methods provide a non-invasive alternative to traditional tools. These methods enhance patient comfort, reduce the risk of infection, and broaden accessibility across healthcare settings. Guan et al. (Guan et al., 2022) demonstrate how Wi-Fi signals can monitor respiratory activity through beamforming and frequency domain analysis. Similarly, Guo et al. (Guo et al., 2022) show that respiration patterns can be used for human identification by analyzing wireless signal variations caused by breathing, leveraging multiple-input-multiple-output (MIMO) technology. The potential applications of contactless methods are broad, ranging from smart healthcare spaces to driver fatigue detection in intelligent vehicles (Gao et al., 2020).

### 2.2. Wi-Fi CSI signals

Channel State Information (CSI) is fine-grained data independently collected by Wi-Fi devices. It characterizes the signal transmission between a transmitter and a receiver, such as a Wi-Fi router and a smartphone or computer. CSI provides detailed amplitude and phase measurements across multiple subcarriers within a Wi-Fi channel. These measurements are sensitive to environmental changes caused by breathing-induced motion (Ma et al., 2019). Fig. 2 illustrates the nature of wireless sensing and its sensitivity to environmental changes such as human respiration.

Breathing causes periodic fluctuations in Wi-Fi signals, and CSI tools capture these fluctuations. Further processing allows for the inference of respiratory rate and patterns. CSI data can be retrieved from off-the-shelf (COTS) Wi-Fi devices on the Wi-Fi protocol stack's physical (PHY) layer using tools such as the Intel 5300 NIC or Atheros CSI Tool (Mosleh et al., 2022; Soto et al., 2022).

Fig. 3 presents a sample of CSI raw data, showing amplitude and phase variations due to multi-path effects.

CSI measurements also capture the impact of signal propagation between multiple antennas in a MIMO Wi-Fi system. Each entry in the CSI matrix represents the channel response between a pair of transmitting and receiving antennas. These matrices are composed of multiple subcarrier measurements, enabling the detection of detailed environmental changes (Soto et al., 2022).

Wi-Fi CSI tools analyze the state information across multiple subcarriers at various frequencies with high granularity, capturing subtle signal variations caused by respiratory movements. The collected CSI data forms a time series of matrices representing the signal propagation characteristics, including amplitude attenuation and phase shifts.

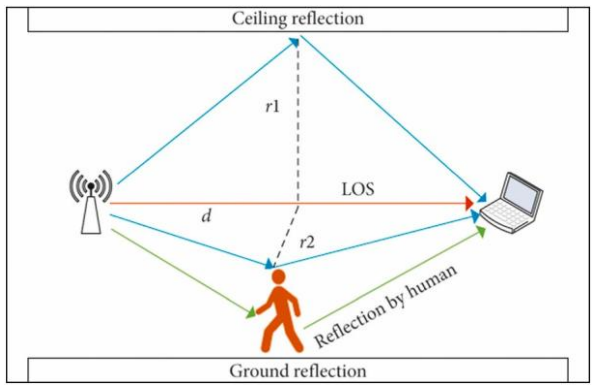

**Fig. 2.** Illustration of Wireless Sensing and Signal Reflection Paths. The figure depicts direct Line of Sight (LOS) transmission and reflections from the ceiling, ground, and a human subject, which are typical in wireless sensing setups. Adapted from (Hao et al., 2020).

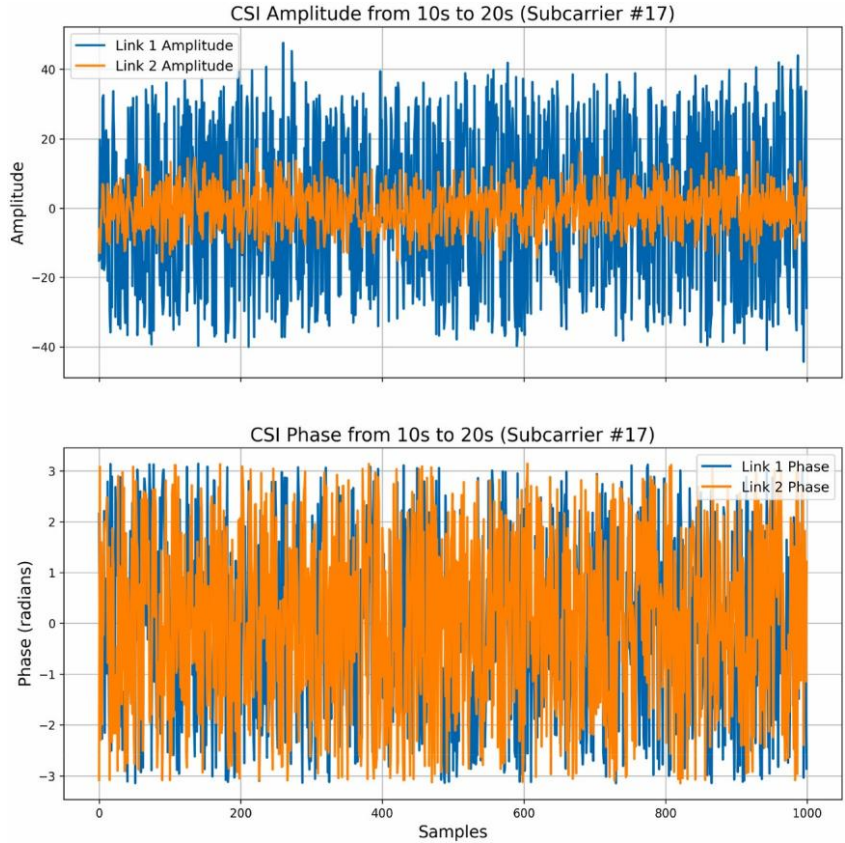

**Fig. 3.** Amplitude and phase variations across selected subcarrier (17) of the CSI data. These changes, sensitive to respiratory movements, illustrate the signal fluctuations that can be analyzed for breath rate detection.

## 2.3. Paper selection approach

We followed a classification-based literature selection method. Specifically, we focused on identifying significant contributions in the fields of application and methodological strategies used in breath analysis.

### 2.3.1. Search strategy

We conducted an extensive literature search using Google Scholar, IEEE Xplore, and Digital Library. The search included keywords such as 'respiratory rate monitoring,' 'Wi-Fi CSI-based breath analysis,' 'acoustic signal-based breath analysis,' 'machine learning for breath analysis,' and 'deep learning for respiratory monitoring.' Prioritizing the highly cited papers or those in top-ranked search results to determine relevance and significance.

### 2.3.2. Scope and selection criteria
Specific criteria guided the selection process to ensure a focused and comprehensive review:

- *Prioritization of Contactless Approaches:* As the survey primarily focuses on contactless methods, particularly Wi-Fi CSI and acoustic sensing, we prioritized papers relevant to these approaches while including selective wearable approach-based studies that were highly cited and offered a strong comparative basis for contactless techniques.
- *Empirical Validation:* Studies incorporating real-world datasets, performance evaluations, or model comparisons were preferred to ensure the reliability and applicability of findings.
- *Time-frame Consideration:* Studies published between 2018 and 2025 were included to reflect recent advancements in breath analysis technologies.
- *Exclusion of Hardware-Centric Studies:* Works focused solely on hardware innovations for wearable sensors, without the integration of machine learning, deep learning, or advanced signal processing, were not considered.

The selected studies were categorized based on their primary area of focus. From an application perspective, the literature was grouped into key domains, including single- and multi-user respiratory rate monitoring. Each category is examined in detail, summarizing the underlying techniques, models employed, and performance evaluations.

From a methodological standpoint, studies were classified into three broad categories: wearable-based, acoustic signal-based, and Wi-Fi CSI-based. The discussion presents a structured comparison of these methods, detailing the data types, sample sizes, and processing techniques used across different studies. Studies within each methodological category are summarized in tabular form to provide a clear overview, offering a systematic analysis of the approaches and their practical implications.

### 2.3.3. Selection justification

While wearable-based breath analysis has been extensively explored in the literature, our selection focused on recent and highly cited studies that provide a comparative foundation for contactless techniques. The chosen work ensures that:

- Applications across both single-user and multi-user scenarios are covered.
- Both traditional (wearable) and evolving (contactless) methods are evaluated.
- Machine learning and deep learning approaches are highlighted.
- Recent trends in non-invasive, real-time breath analysis are emphasized.

This survey presents a well-organized comparative review of the state-of-the-art in-breath analysis by structuring our selection along both application domains and methodological approaches.

## 3. Applications

Breath analysis systems have many applications that significantly advance healthcare, security, and personal monitoring. These applications leverage advanced technologies to offer accurate, non-invasive, and real-time monitoring of respiratory activities. This section discusses the key applications, highlighting their significance and contributions to respiratory analysis for healthcare monitoring.

**Table 2**
Summary of single-user respiratory rate detection studies.

| Reference | Technique | Model | Performance/Accuracy |
|-----------|-----------|-------|----------------------|
| Xu, Guo, and Chen (2020) | Combine Peak and Frequency Domain | Not specified | <0.9 BPM (standing), <0.4 BPM (sitting) |
| Brieva et al. (2023) | Hermite Transform to implement EMT | CNN and AHN | CNN: 78.97 %, AHN: 97.22 % |
| Hu et al. (2022) | Deep CNN | CNN | 96.05 % (single user) |
| Rehman et al. (2021) | Curve fitting and ML algorithms | Curve fitting and ML algorithms | Varied accuracy based on the model |
| Park et al. (2023) | Near-infrared spectroscopy (NIRS) and CNN | Pre-ResNet-based 1D CNNs with 1 × 3 and 1 × 5 Stage 1 conv layers for DR | 88.79 % (without Stage 1), 90.58 % (1 × 3 Stage 1), 91.77 % (1 × 5 Stage 1) |
| Islam et al. (2018) | 4-channel lung sound data ac-question | SVM, ANN | SVM: 70.0 %, 76.7 %, 73.3 %; ANN: 75.9 %, 78.1 %, 77.0 % |
| Tarim et al. (2023) | Accelerometer and temperature sensor for breath analysis, YOLOv5m for classification | YOLOv5m trained for inhale, exhale, and breathlessness detection | Accuracy: 96 %, Sensitivity: 97.6 %, Specificity: 79.7 % |
| Yin et al. (2021) | Smartphone-based Wi-Fi signals | Smartphone-based system | High accuracy for respiration sensing |
| Srivastava et al. (2021) | MFCC, Mel-Spectrogram, Chroma with CNN for COPD detection | CNN using Librosa with 10fold Cross-Validation | ICBHI score of 93 % |
| Romano et al. (2023) | Microphone-based breath sound analysis | Bandpass filtering (200–800 Hz), Hilbert transform for envelope extraction | MAE: 1.7 bpm (all conditions), 3.8 bpm max (running at 12 km/h) |
| Srivastava et al. (2021) | Deep Learning Respiratory rate prediction from bio-signals | LSTM, Bi-LSTM, CNN-LSTM, Attention- based LSTM Bi-LSTM with Bahdanau attention: | MAE = 0.24 ± 0.03 (PPG and ECG), 0.51 ± 0.03 (sEMG) |
| Guo et al. (2023) | Multi-Antenna CSI, ICA, Gaussian-HMM | MCG-HMM | Mean absolute error 0.1 BPM |
| Fan, Yang, et al. (2024) | CNN-LSTM Model | CNN-LSTM | Accuracy: 97.8 % (test set), up to 98.1 % in different environments |
| Gui et al. (2022) | CNN-based turnover recognition, Subcarrier selection for breath rate | CNN | Turnover recognition: 94.59 %. Breath rate estimation improved in prone/side-lying positions |
| Ali, Alloulah, et al. (2021) | First-order Differentiation, PCA-based Motion Tracking | PCA-based CSI Processing | Average breath rate error: <1.19 BPM |
| Lalouani et al. (2022) | Acoustic segmentation | Breath cycle segmentation & scoring | Effective in identifying respiratory anomalies but sensitive to noise |
| Chara et al. (2023) | FMCW sonar | Hilbert Transform, peak detection | Accuracy <0.15 BPM error |
| Hou et al. (2022) | Microphone array processing | Beamforming, noise filtering | 12 % improvement over existing systems |
| Xu, Yu, and Chen (2020) | Smartphone acoustic tracking for Single-User Respiratory Monitoring (Driver safety) | ESD, GAN | 0.11 BPM error, 0.95 correlation with ground truth |

### 3.1. Single-user respiratory rate detection

Single-user respiratory rate detection is fundamental for personal health monitoring and diagnostic applications. Accurate and reliable respiratory rate measurements are crucial for individual health assessment and early disease detection (Addison et al., 2023).

Researchers have proposed diverse techniques to measure single-user respiratory activities, including Wi-Fi sensing, smartphone- based sensing, and acoustic signal sensing. These methods leverage unique characteristics of wireless signals and advanced computational techniques to capture and analyze respiratory patterns accurately. Wi-Fi CSI, for example, can detect minute variations in signal amplitudes and phases caused by respiratory movements, enabling non-invasive monitoring. Smartphone-based systems utilize the ubiquitous nature of mobile devices to offer convenient and accessible respiratory rate detection. Deep learning algorithms, such as Long Short-Term Memory (LSTM) networks, are then employed to process complex signal patterns and improve the accuracy of respiratory rate measurements.

Table 2 summarizes the techniques and performance of some single-user respiratory rate detection studies, highlighting the models used and their respective accuracy.

Deep learning approaches have consistently demonstrated greater accuracy in respiratory rate estimation. Hu et al. (Hu et al., 2022) employed a deep CNN with 96.05 % accuracy, while Fan et al. (Fan, Yang, et al., 2024) employed a CNN-LSTM framework to record 97.8 % accuracy in diverse environments. Srivastava et al. (Srivastava et al., 2021) employed CNN-LSTM models to detect COPD and recorded an ICBHI score of 93 %. These findings indicate that deep learning frameworks effectively encapsulate complex respiratory patterns and are highly suited to applications requiring high accuracy.

Although less computationally intensive, machine learning techniques can be as competitive in accuracy when combined with well- engineered feature extraction. Islam et al. (Islam et al., 2018) used SVM and ANN models to classify lung sounds, and ANN achieved 78.1 % accuracy. Similarly, Rehman et al. (Rehman et al., 2021) utilized a combination of curve fitting and machine learning algorithms and achieved varying degrees of accuracy based on the model chosen. These methods are interesting since they take less computation and offer sufficient accuracy.

Traditional signal processing remains relevant when prioritizing real-time computation over complex model training. Xu et al. (Xu, Guo, & Chen, 2020) applied frequency domain analysis and peak detection, achieving a mean absolute error (MAE) of less than 0.9 BPM in standing positions and 0.4 BPM in sitting. Romano et al. (Romano et al., 2023) used Hilbert Transform and bandpass filtering, obtaining an MAE of 1.7 BPM across different conditions. Although these methods require careful noise reduction techniques, they provide computational efficiency suitable for resource-constrained environments.

### 3.2. Multi-user respiratory rate detection

While single-user respiratory rate detection systems are practical for personal health monitoring, they face limitations in crowded environments with multiple users. Multi-user respiratory rate detection is necessary for applications in public spaces and other overcrowded environments where distinguishing between individual respiratory signals is essential.

Various works have explored how to detect multi-user respiratory activities, mainly in lab settings. Researchers have employed techniques such as CSI amplitude sensing, beamforming, and clustering algorithms to differentiate and accurately measure respiratory rates from multiple users. These methods use advanced signal processing and machine learning techniques to resolve the challenges of overlapping signals and interference. For example, orthogonal frequency-division multiplexing (OFDM) and fast Fourier transform (FFT) are used for signal pattern analysis in complex environments. In contrast, clustering algorithms like density-based spatial clustering of applications with noise (DBSCAN) are used in separating and identifying individual respiratory signals.

Table 3 summarizes the techniques and performance of several multi-user respiratory rate detection studies, highlighting the models used and their respective performances; the study shows that Deep learning models have demonstrated strong performance in multi-user scenarios by leveraging advanced signal processing and feature extraction. Ali et al. (Ali, Alloulah, et al., 2021) combined Doppler spectral analysis with CNNs, achieving 98 % accuracy in people counting with a root mean square error (RMSE) of 0.13 BPM.

**Table 3**
Summary of multi-user respiratory rate detection studies.

| Reference | Technique | Model | Performance/Accuracy |
|---|---|---|---|
| Wang et al. (2023) | Dualforming for SSNR enhancement with MUS2IC | Dualforming with MUS2IC | <1.29 BPM at 120° (60° each side) |
| Wan et al. (2023) | CSI amplitude sensing | OFDM and FFT | <1 BPM for multi-user moving scenario |
| Guan et al. (2022) | Beamforming and DBSCAN | DBSCAN | 97.04 % accuracy in multi-user scenarios |
| Gao et al. (2020) | Clustering with DAM | Clustering with DAM | 97.5 % in 3 users |
| Zhang et al. (2021) | LoRa CSI with beamforming | Beamforming | 98.1 % accuracy in five human subjects |
| Xiong et al. (2020) | SIMO radar with adaptive digital beamforming (ADBF) | ADBF with m-Capon and LCMV | High detection accuracy for multi-target scenarios |
| Liang et al. (2021) | Spectroscopy, CE-DFCS | CE-DFCS | High Sensitivity |
| Ali, Alloulah, et al. (2021) | CIR, Breathing-to-Noise Ratio (BNR), Multi-Sense ICA | Multi-Sense ICA | 0.61 BPM error (multi-user scenario) |
| Ali, Alloulah, et al. (2021) | Doppler Frequency Analysis, Subcarrier Selection | Doppler Spectral Analysis | 98 % people-counting accuracy, 0.13 BPM RMSE |
| Fan, Pan, et al. (2024) | Short-Time Window Shift, Multi-Antenna CSI Processing | RespEnh Framework | 80 % detection rate within 1 BPM error at 6.4m |

Fan et al. (Fan, Pan, et al., 2024) introduced the RespEnh framework, which achieved an 80 % detection rate within 1 BPM error at 6.4 m. These results highlight the ability of deep learning models to generalize well in complex environments.

Machine learning methods such as clustering and independent component analysis (ICA) have also been successful in multi-user respiratory detection. Guan et al. (Guan et al., 2022) employed DBSCAN clustering to separate individual respiratory patterns, achieving 97.04 % accuracy. Liang et al. (Liang et al., 2021) leveraged CE-DFCS spectroscopy, demonstrating high sensitivity in detecting multiple users. These methods provide a robust alternative to deep learning by reducing computational complexity while maintaining high accuracy.

Traditional signal processing techniques, particularly beamforming and frequency analysis, remain helpful in structured environments with predefined spatial separation. Wang et al. (Wang et al., 2023) utilized dual-forming with MUS2IC, achieving an error of less than 1.29 BPM at 120-degree separation angles. Wan et al. (Wan et al., 2023) applied FFT-based CSI amplitude sensing, achieving an error of less than 1 BPM in mobile multi-user scenarios. These results suggest that traditional techniques perform well in controlled setups but may struggle in highly dynamic environments.

While deep learning and clustering-based machine learning approaches provide superior accuracy, traditional frequency-domain techniques remain viable for constrained environments where real-time processing is required.

### 3.3. User identification

User identification through breath patterns presents a novel and secure approach to biometric authentication. An individual's unique breathing pattern provides a noninvasive method for user authentication, which can be critical for security systems (Guo et al., 2022).

Fig. 4 provides a high-level overview of the Breath ID system. The process begins with data collection from various devices, including Wi-Fi-based systems, IoT devices with acoustic sensors, and wearable devices. These devices gather multiple data types, such as CSI phase and amplitudes, Doppler information, acoustic signals, etc. The collected data undergoes preprocessing steps and analysis using signal processing, machine learning, or deep learning techniques. The final output is identifying users based on their unique respiratory signatures.

The studies summarized in Table 4 demonstrate the viability of using breath-related signals for identity recognition. Several approaches stand out in terms of accuracy, robustness, and practicality.

Deep learning methods have shown strong performance in user identification. For instance, Wang et al. (Wang et al., 2021) implemented a three-layer CNN that achieved 98.3 % accuracy in non-interference conditions. Similarly, Pham et al. (Pham et al., 2022) applied CNN-LSTM models to multimodal sensor data, attaining 95 % accuracy in one-shot identification and above 90 % across breathing variations. These results demonstrate the adaptability and precision of deep learning models in recognizing subtle biometric features from respiration.

Machine learning classifiers provide competitive performance with reduced computational complexity. Guo et al. (Guo et al., 2022) employed a WMD-DTW algorithm on CSI data to achieve 97.5 % identification accuracy, making it a suitable option for lightweight authentication systems. Other approaches, like clustering methods (Guan et al., 2022) and spectral filtering (Wang, Zhang, et al., 2020), have also demonstrated promising results in multi-user scenarios.

Although traditional signal processing techniques are less prevalent, they remain relevant for resource-constrained or low-power applications. Wang et al. (Wang, Zhang, et al., 2020) utilized FFT and iterative dynamic programming to achieve over 85 % accuracy. Template-matching methods, such as the one proposed by Lu et al. (Lu et al., 2017), offer low error rates and high robustness under stable environmental conditions.

In summary, breath-based user identification systems benefit significantly from deep learning architectures regarding accuracy and scalability. However, simpler models such as DTW-based or template-matching classifiers still offer meaningful trade-offs where interpretability, real-time computation, or energy efficiency are critical.

### 3.4. Heartbeat monitoring

Breath Analysis for Heartbeat monitoring is essential for personal health management and clinical settings. Accurate monitoring of heartbeat rates can lead to early detection of cardiovascular issues, enhance patient care, and support continuous health management for individuals with chronic heart conditions. This is essential in single-user environments, such as personal health monitoring at home, and multi-user environments, like hospitals and elder care facilities, where multiple individuals need simultaneous monitoring. Accurate heartbeat monitoring faces significant challenges due to interference and overlapping signals, especially in multi-user environments. However, advanced sensing technologies have been developed to address these challenges and provide reliable data.

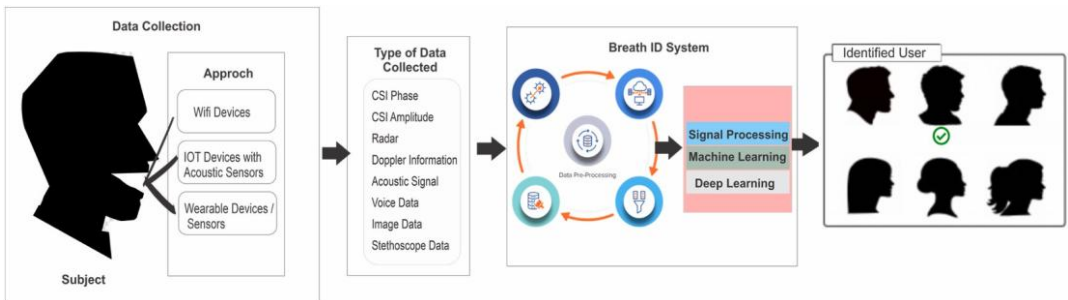

**Fig. 4.** User identification applications structure.

**Table 4**
Summary of user identification studies.

| Reference | Technique | Model | Performance/Accuracy |
|---|---|---|---|
| Guo et al. (2022) | CSI sensing with weighted matching divergence (WMD) and dynamic time- warping (DTW) | WMD-DTW algorithms | 97.5 % accuracy |
| Aslam et al. (2021) | Wi-Fi CSI with machine learning for activity pattern detection | Machine learning model | High sensitivity to activity nuances |
| Guan et al. (2022) | Clustering respiratory signals using Density- based Adaptive Matching (DAM) | DAM Clustering | 97.5 % accuracy in a 3-user setup |
| Pham et al. (2022) | IoT device with multi-modal data (acoustic sensor, accelerometer, gyroscope) | CNN-LSTM, TCN | Multi-modal: 95 % accuracy (K = 1), 93–94 % for varied breaths; Mono-modal: Audio 86 %, Motion 88 % |
| Zhao et al. (2017) | I-vector and constant-Q spectrogram for feature extraction | CNN-LSTM | 91.3 % accuracy in speaker identification |
| Tran and Tsai (2019) | Stethoscope-acquired data vs traditional microphone data | Signal Comparison Model | High accuracy in low-noise environments |
| Tran et al. (2023) | Mobile device-acquired bronchial breath sound data with data augmentation and SSL | Breath PID's SSL Model | Effective identification based on high-frequency components |
| Lu et al. (2017) | MFCC-based template matching for respiratory patterns | Template matching (BreathID) | 0.04 % FIR, 0.12 % FAR, 0.15 % FRR |
| Liu et al. (2020) | CSI extraction from Wi-Fi signals with fuzzy wavelet transformation | Deep Neural Network (DNN) | Verification: >95 %, Spoofing detection: 92 %, False Positive Rate: <5 % |
| Wang, Zhang, et al. (2020) | Wi-Fi CSI-based respiration analysis for person identifications | FFT, Spectral filtering, Iterative Dynamic Programming (IDP) and Markov Chain Model | Accuracy: People counting (86 %), recognition (85.78 %) |
| Wang et al. (2021) | CNN, Weighted Subcarrier Screening, Feature Integration | Three-layer CNN | 98.3 % accuracy (non-interference), 91.2 % (with interference) |

The studies summarized in Table 5 demonstrate the growing capability of using breath-related sensing for accurate heartbeat monitoring. Signal processing methods are the most widely adopted, particularly in radar and Wi-Fi-based systems, due to their robustness and ease of deployment.

Frequency-domain techniques, such as those based on FFT and wavelet transforms, have shown strong performance. Zhang et al. (Zhang et al., 2020) used frequency-modulated continuous wave (FMCW) sensing and FFT to achieve a heart rate estimation error as low as 0.75 BPM using smart speakers. Similarly, Wang et al. (Wang, Yang, & Mao, 2020) demonstrated effective heartbeat extraction through Wi-Fi CSI signals by combining subcarrier selection with a discrete wavelet transform, achieving high precision in both heart and breath rate estimation.

Machine learning-based techniques have also emerged in heartbeat pattern recognition. McClure et al. (McClure et al., 2020) applied convolutional neural networks (CNNs) to classify heartbeat patterns under various breathing conditions. While not as prevalent as in other applications like user identification, deep learning models are gaining traction due to their potential to improve robustness and feature extraction in noisy environments.

In summary, heartbeat monitoring through breath analysis is a promising area where frequency-domain signal processing remains dominant. However, introducing machine learning and deep learning models, particularly CNNs, leads to a shift toward more adaptive and intelligent systems. Future work will likely involve refining deep learning architectures and exploring hybrid models that integrate

**Table 5**
Summary of heartbeat monitoring studies.

| Reference | Technique | Model | Performance/Accuracy |
|---|---|---|---|
| Wang et al. (2023) | SSNR enhancement with dual forming for multi-user monitoring | Dual-forming | <1.29 BPM at 120° (60° each side) |
| Wang, Yang, and Mao (2020) | Wi-Fi CSI with data calibration, subcarrier selection, and discrete wavelet transform (DWT) | Signal Processing Pipeline | High accuracy for heartbeat and respiration detection |
| Zhang et al. (2020) | Acoustic sensing using Frequency Modulated Continuous Wave (FMCW) and FFT for precise monitoring | FMCW and FFT | Heart rate estimation error of 0.75 bpm |
| Wang et al. (2017) | Continuous-wave (CW) radar with OFDM Doppler shift | OFDM-based Processing | Enhanced accuracy and reduced complexity |
| McClure et al. (2020) | CNN-based classification for pattern recognition | Convolutional Neural Network (CNN) | Varies across different breathing types |
| (Gouveia et al. (2024)) | Motion artifact reduction in PPG for SpO2 and pulse rate | OxiMA: Frequency-domain reconstruction | PR error: 3 bpm, SpO2 error: 3.24 % |

traditional and data-driven methods to enhance performance across diverse environments.

*3.5. Disease detection and monitoring applications*

Detecting and monitoring respiratory diseases through non-invasive techniques significantly enhances diagnostic accuracy and patient comfort, providing critical tools in clinical and home healthcare settings. Various studies have developed novel approaches to monitor respiratory patterns for non-contact breath analysis, which is essential in early disease detection, chronic illness management, and continuous health monitoring, particularly for conditions like COVID-19 and COPD.

The studies summarized in Table 6 demonstrate the growing potential of breath-based sensing technologies for disease detection and health monitoring. Various modeling techniques have been employed across deep learning, machine learning, and traditional signal processing paradigms, offering distinct accuracy, scalability, and deployment feasibility advantages.

Deep learning approaches have shown excellent performance in extracting complex respiratory features for disease detection. Gong et al. (Gong et al., 2022) applied CNN-based feature extraction to acoustic sensing data, achieving high accuracy through transfer learning and user-independent validation. Similarly, Moshiri et al. (Moshiri et al., 2021) utilized both 2D CNN and BLSTM models on CSI signals, with CNN achieving 95.5 % accuracy. Kumar et al. (Kumar et al., 2021) analyzed cough spectrograms using deep neural networks, while Wang et al. (Wang, Hu, et al., 2020) used a BI-AT-GRU network to classify six respiratory patterns with an accuracy of 94.5 %. These studies underscore the potential of deep architectures in developing robust, generalized, and high-performing disease monitoring systems.

**Table 6**
Summary of disease detection and health monitoring studies.

| Reference | Technique | Model | Performance/Accuracy |
|---|---|---|---|
| Gong et al. (2022) | Acoustic sensing combined with CNN-based feature extraction | Convolutional Neural Networks (CNN) | Validated with leave-one-user out cross-validation and transfer learning, accuracy is high. |
| Moshiri et al. (2021) | CSI data processing using Raspberry Pi with 2D CNN and BLSTM models | 2D CNN and BLSTM | 2D CNN: 95.5 %, BLSTM: 94.7 % |
| Dou and Huan (2021) | IFFT and SHFT for signal processing | IFFT and SHFT | Error <0.7 BPM, Avg error: 0.15 BPM |
| Nallanthighal et al. (2021) | Fusion-based approach with temporal-spatial features | CNN and LSTM models | Fusion approach enhances the accuracy |
| Fraiwan et al. (2021) | Ensemble classification on stethoscopic lung sounds | Ensemble classifiers | High accuracy across multiple classifiers |
| Purnomo et al. (2021) | FMCW radar for non-contact respiratory pattern detection | FMCW radar system | Reliable accuracy in pattern detection |
| Hernandez and Bulut (2020a) | IoT system for remote health applications | IoT-based framework | Improved healthcare monitoring efficiency |
| Yue et al. (2018) | IoT-enabled deep learning for respiratory monitoring | Deep learning models | High accuracy in monitoring |
| Wang et al. (2017) | Secure IoT architecture for health monitoring | IoT security framework | Enhanced security and scalability |
| Kumar et al. (2021) | Deep learning on cough spectrograms | Deep neural network | Effective disease detection from coughing patterns |
| Srivastava et al. (2021) | ML-based audio analysis for respiratory anomalies | Machine learning models | High accuracy in anomaly de-tection |
| Xu et al. (2019) | ML applied to breathing sounds for health monitoring | Machine learning-based analysis | High accuracy under varying conditions |
| Sait et al. (2021) | Wi-Fi signal processing for respiratory monitoring | Novel ML/DL methods | Improved monitoring and signal processing accuracy |
| Wang, Hu, et al. (2020) | Depth camera with BI-AT-GRU for respiratory health monitoring | BI-AT-GRU model | Achieved 94.5 % accuracy |
| Molin et al. (2021) | Obstructive sleep apnea (OSA) prediction using smartphone recordings | FFT + Random Forest, SVM, Naïve Bayes, LDA | RF: 93 %, SVM: 90 %, Manual interpretation: 55 % |
| Indrawati et al. (2022) | Obstructive sleep apnea detection | ECG-based FFT + ANN, SVM, KNN, LDA | ANN: Accuracy = 84.64 %, Sensitivity = 94.21 %, Specificity = 64.03 % |
| Blanco-Almazan et al. ´ (2021) | COPD respiratory phase detection | Bioimpedance-based Respiratory Phase Detection Algorithm | Accuracy: 96–99 %, Mean Absolute Percentage Error (MAPE) < 3.42 %. |
| Alzaabi et al. (2024) | ESP32-based CSI Monitoring, Bland-Altman Analysis | Bland-Altman Method | Validity: [1.29, 1.06] BPM error, 80 % repeatability at 14 BPM |
| Khan et al. (2023) | PCA, SVM, EMD, Fine Tree Algorithm | Fine Tree Algorithm | 87.5 % BPM estimation, 96.9 % (multi- class), 95.8 % (binary class) |
| Verde et al. (2023) | Voice feature extraction (MFCC, spectral features) for COVID-19 Detection | Random Forest, SVM, CNN | Accuracy 85–92 % (depending on vowel sounds) |
| Shimizu et al. (2023) | Disease Detection and Monitoring (Aspiration risk in anesthesia) using AI-based Stridor Quantitative Value (STQV) | AI-based anomaly detection | Detects 87 % of breath sounds during capnography apneas |
| Skalicky et al. (2021) | Breath phase segmentation for Lung disease monitoring | High-pass & low-pass filtering, envelope detection | Achieves High specificity |

Machine learning models continue to provide practical solutions, mainly where datasets are limited or interpretability is a concern. Fraiwan et al. (Fraiwan et al., 2021) employed ensemble classifiers to analyze lung sounds from stethoscopic recordings, achieving high diagnostic accuracy across various models. Verde et al. (Verde et al., 2023) used voice features and tested several ML algorithms, including random forest, SVM, and CNN—for COVID-19 detection, reporting up to 92 % accuracy. Molin et al. (Molin et al., 2021) and Indrawati et al. (Indrawati et al., 2022) utilized traditional classifiers such as SVM, KNN, and ANN for sleep apnea detection, showing promising results in both sensitivity and specificity.

Signal processing techniques remain crucial, especially in resource-constrained environments where model complexity and real- time computation are limiting factors. Dou et al. (Dou & Huan, 2021) applied IFFT and SHFT to extract respiratory features, achieving a low mean error of 0.15 BPM. Purnomo et al. (Purnomo et al., 2021) used FMCW radar to monitor respiratory activity with reliable accuracy, emphasizing the effectiveness of radar-based systems for non-contact applications. Xu et al. (Xu et al., 2019) and Skalicky et al. (Skalicky et al., 2021) leveraged feature extraction and filtering methods for breath phase segmentation and anomaly detection in disease monitoring tasks.

In summary, deep learning models offer the highest diagnostic performance and generalization capacity, making them suitable for scalable and automated disease monitoring solutions. Machine learning approaches remain advantageous for lightweight systems with constrained data and interpretability needs. Signal processing remains highly relevant for real-time and low-cost monitoring, especially with emerging sensing modalities like Wi-Fi CSI and radar. Together, these approaches represent a comprehensive toolkit for advancing non-invasive respiratory disease detection in clinical and at-home settings.

Overall, deep learning consistently provides the highest accuracy in most applications at the expense of tremendous computation costs. However, machine learning methods achieve a perfect trade-off between efficiency and accuracy. In contrast, conventional signal processing methods remain viable for real-time low-power applications, where computational efficiency is essential.

## 4. Approaches/Methods

This section reviews the various approaches and methods employed in breath analysis. We begin with wearable device approaches, discussing their concepts, advantages, and limitations. Subsequently, we explore contactless approaches, highlighting the benefits of acoustic signal processing and Wi-Fi CSI methods. These evolving methods address the challenges posed by wearable devices, enhancing user comfort, minimizing risk, and advancing data quality.

### 4.1. Contact-based (wearable device) approach

Wearable technology has traditionally been a foundational tool for continuously monitoring respiratory parameters. These devices typically incorporate biosensors, accelerometers, and gyroscopes—to record chest movements and other physiological signals associated with breathing. By positioning these sensors around the chest or abdomen, continuous data can be gathered, allowing for real- time tracking and analysis of respiratory patterns. This setup provides critical insights into respiratory health, making wearable devices invaluable in chronic condition management and patient monitoring.

**Table 7**
Summary of wearable devices approaches.

| Reference | Approach | Data Type | Data Size | Data Processing Method |
|---|---|---|---|---|
| McClure et al. (2020) | Wearable sensor | Accelerometer and gyroscopic data | Samples from 100 healthy subjects | CNN for feature extraction |
| Park et al. (2023) | Deep learning for breath pattern classification | Tissue hydrodynamic response | From 11 subjects | Data size reduction with CNN |
| Islam et al. (2018) | Lung sound analysis | Posterior lung sound signals | 4-channel data acquisition | SVM classifier and ANN classifier |
| Tarim et al. (2023) | Deep learning for wearable health monitoring | Wearable sensor data | 100+ subjects | Captured using wearable sensors; pre- processed with signal filtering and segmentation; processed with CNNs and LSTMs |
| Pham et al. (2022) | Multimodal fusion | Audio, Accelerometer, Gyroscope | 20 subjects, multiple sessions | Annotation using Audacity tool and processed using CNNLSTM, TCN |
| Tran and Tsai (2019) | Comparison of traditional and stethoscope acquired data | Audio recordings from stethoscope and microphone | Not specified | Noise reduction, signal enhancement, Comparison analysis |
| Tran et al. (2023) | Recording bronchial breath sounds at optimal neck positions | Bronchial breath sounds | Not specified | Selection of optimal recording positions; Frequency analysis |
| Blanco-Almazan ´ et al. (2021) | Algorithms to detect respiratory phases | Bioimpedance and airflow | 50 COPD patients | Low-pass filtering, up-sampling, and zero- crossing detection |
| Kumar et al. (2022) | Wearable (ECG, PPG, sEMG-based) | ECG, PPG, sEMG Capnobase, MIMIC-II, | Diaphragm sEMG datasets | FFT, feature extraction, MAEbased evaluation |
| Harvey et al. (2018) | Wearable (PPG-based Signal reconstruction) | PPG signals (photoplethysmography) | 10 subjects (22min sessions) | Time-frequency decomposition, spectral filtering |
| Indrawati et al. (2022) | Wearable (ECG derived frequency analysis) | ECG RR intervals | MIT-BIH Apnea database | ANN, FFT, entropy, geometric mean, statistical feature extraction |

**Advantages of the Wearable Devices Approach**

- Wearable devices are widely used in various health monitoring applications, and many users are already familiar with this technology, making it easier to adopt.
- These devices offer continuous monitoring capability, providing valuable data for chronic respiratory conditions and other long- term health issues.
- When adequately calibrated, wearable devices can provide accurate measurements of respiratory parameters, contributing to reliable health assessments.
- The variety of sensors embedded in wearable devices allows for the collection of multiple physiological signals simultaneously, providing a comprehensive view of the user's respiratory health.

**Limitations of the Wearable Devices Approach**

- Continuous use of wearable devices can be uncomfortable and inconvenient for users, particularly for long-term monitoring.
- Wearable devices require regular maintenance and cleaning, burdening users.
- Wearable sensors typically monitor a limited area, which may not provide comprehensive respiratory data.
- Movement and environmental factors can introduce noise and artifacts in the collected data, affecting accuracy.

Table 7 summarizes key wearable device-based approaches for respiratory monitoring. These approaches employ a range of techniques— from deep learning classifiers to multimodal sensor fusion—illustrating the versatility of wearable technology in detecting respiratory patterns. For example, Park et al. (Park et al., 2023) utilized tissue hydrodynamic response data from wearable devices to classify breath patterns using a 1D CNN, demonstrating high accuracy but facing the challenge of user discomfort during prolonged use. In another study, Islam et al. (Islam et al., 2018) employed a 4-channel lung sound analysis system to classify normal and asthmatic subjects, achieving higher accuracy with ANN than with SVM classifiers. This underscores the effectiveness of deep learning models for biosensor data processing in respiratory disease classification. Additionally, Blanco et al. (Blanco-Almazan et al., ´ 2021) utilized bioimpedance and airflow measurements to identify respiratory phases in COPD patients, applying low-pass filtering, up-sampling, and zero-crossing detection with high accuracy and low Mean Average Percentage Error (MAPE).

While wearable devices offer valuable benefits in continuous monitoring and data richness, they also pose challenges regarding user comfort and potential data accuracy issues due to movement and environmental factors. To mitigate these limitations, contactless methods have emerged as a promising alternative, leveraging technologies like acoustic signal processing and Wi-Fi CSI.

*4.2. Contactless approaches*

Contactless approaches to breath analysis offer non-invasive alternatives to traditional wearable devices by capturing respiratory signals without direct physical contact. This can be achieved through acoustic signals and Wi-Fi Channel State Information (CSI), which enhance user comfort, minimize infection transmission risk, and improve data accuracy. By analyzing audio-based signals or wireless transmission variations, these approaches enable respiratory monitoring without physical interaction. This section explores two prominent contactless methods: acoustic signal processing and Wi-Fi CSI.

*4.2.1. Acoustic Signal Approach*

Acoustic signal processing is particularly effective in environments with minimal external noise interference, such as homes, sleep studies, and healthcare facilities. This approach uses microphones or acoustic sensors near the subject to capture breath sounds, which are processed by algorithms analyzing frequency and amplitude variations to assess respiratory parameters. Acoustic signal processing provides a noninvasive, continuous monitoring solution for respiratory analysis by distinguishing between different breathing patterns.

**Advantages of Acoustic Signal Approach**

- Non-contact method enhances comfort and is suitable for long-term use.
- Simple installation and integration with environments, particularly home and healthcare settings.
- Continuous monitoring capabilities are ideal for applications requiring ongoing health assessment.

**Limitations of Acoustic Signal Processing**

- Susceptible to environmental noise interference, which can degrade accuracy.
- Limited effective range, restricting use in larger spaces or high-noise environments.
- Privacy concerns due to the use of microphones, especially in sensitive environments.

Fig. 5 shows the general acoustic signal processing approach, where sound waves generated during respiration are captured by a microphone array near the subject. These signals undergo noise reduction, transformation, and feature extraction, which enables the detection and analysis of respiratory patterns. This technique is applicable in single-user and multi-user breath monitoring across various settings.

Table 8 summarizes studies using acoustic signal processing for breath analysis, showcasing techniques, and data processing methods that enhance respiratory monitoring accuracy. For example, Gong et al. (Gong et al., 2022) introduced BreathMentor, which uses acoustic sensing with CNNs to accurately classify breathing patterns and calculate the inspiratory/expiratory ratio. Wang et al. (Wang et al., 2023) designed an acoustic monitoring system that leverages dual forming to enhance SSNR, achieving precision in multi-user monitoring within home settings. These approaches exemplify the adaptability of acoustic signal processing in healthcare applications, where non-invasive and continuous respiratory assessment is increasingly essential.

### 4.2.2. Wi-Fi using Channel State Information (CSI)

Wi-Fi CSI-based methods leverage the phase and amplitude of Wi-Fi signals to detect respiratory patterns, as explained in Section 2. These methods are particularly advantageous due to their non-invasive nature, broad area coverage, and robustness against environmental noise. By analyzing the changes in Wi-Fi signal properties caused by breathing, Wi-Fi routers and receivers can monitor respiratory patterns without requiring physical contact. Fig. 6 illustrates the general Wi-Fi CSI process for breath analysis. The diagram showcases how Wi-Fi signals interact with the human body during respiration, causing signal amplitude and phase variations. These variations are captured by Wi-Fi devices and CSI Tools, which then process the signals to detect and analyze respiratory patterns. The figure also highlights how the CSI signal undergoes a series of stages to extract respiratory patterns, which are then processed. It is a versatile tool for single-user and multi-user monitoring applications in various environments.

**Advantages of Wi-Fi CSI:**

- **Enhanced Accuracy:** Wi-Fi CSI methods provide higher accuracy in detecting respiratory patterns due to their ability to capture fine-grained signal variations.
- **Robustness to Noise:** Wi-Fi signals are less susceptible to environmental noise than acoustic signals, ensuring more reliable respiratory monitoring.
- **Broad Area Coverage:** Wi-Fi CSI can cover larger areas, making it suitable for simultaneously monitoring multiple users in different locations.
- **Seamless Integration:** Wi-Fi CSI systems can be integrated with existing Wi-Fi infrastructure, reducing the need for additional hardware.

**Limitations of Wi-Fi CSI:**

- **Complexity of Setup:** Setting up and calibrating Wi-Fi CSI systems can be complex and require specialized knowledge.
- **Privacy Concerns:** Like acoustic sensing, Wi-Fi CSI may raise privacy concerns as it involves monitoring through wireless signals.
- **Dependency on Wi-Fi Infrastructure:** The effectiveness of Wi-Fi CSI depends on the quality and availability of the existing Wi-Fi infrastructure, which can vary across different locations.

Table 9 summarizes different Wi-Fi CSI approaches used in the respiratory analysis, highlighting their methods, data types, and processing techniques, showcasing their versatility and robustness. For example, Zeng et al. (Zeng et al., 2018) utilized the complementarity of CSI phase and amplitude for respiration detection. The method employs Fast Fourier Transform (FFT) and Savitzky-Golay

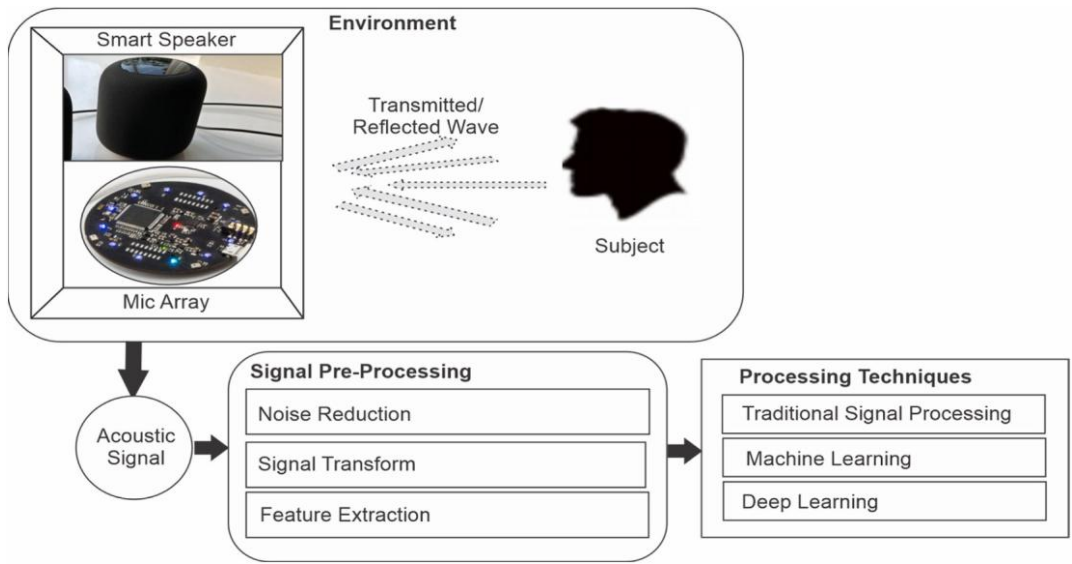

**Fig. 5.** Illustration of acoustic signal processing for breath analysis.

**Table 8**
Summary of acoustic signal processing approaches.

| Reference | Approach | Data Type | Data Size | Data Processing Method |
|---|---|---|---|---|
| Gong et al. (2022) | Acoustic sensing and CNN | Acoustic signals from smart speakers and mic arrays | 31,230 breathing cycles | Convolutional Neural Networks (CNN) for feature extraction |
| Wang et al. (2023) | SSNR enhancement with dual forming | Acoustic signals from smart speakers and mic arrays | Not specified | Dualforming, EMD, and ICEE-MAN for processing |
| Zhao et al. (2017) | Feature extraction using i-vectors and constant-Q spectrograms | Audio recordings | 3000+ instances from 50 speakers | Phoneme segmentation, breath sound extraction, and CNNLSTM |
| Lu et al. (2017) | MFCC-based template matching | Audio recordings | Collected from 50 users | Normalization, feature extraction, and template matching |
| Molin et al. (2021) | Acoustic (Smartphone-based breath sound analysis) | Smartphone-recorded breath amplitudes | Collected from 42 subjects | FFT, feature selection, multiclass classification |
| Romano et al. (2023) | Acoustic (Microphone in wearable facemask) | Audio recordings (breath sounds) | Collected from 10 subjects | Bandpass filtering (200–800 Hz), Hilbert transform for envelope extraction. |
| Lalouani et al. (2022) | Acoustic breath segmentation for disease detection | Breath sound recordings from COPD patients | Public dataset | Spectrogram analysis, breath phase segmentation |
| Chara et al. (2023) | Smartphone-based acoustic respiration tracking | Sonar-based breathing signals | Experimental setup | Hilbert Transform, noise removal |
| Hou et al. (2023) | Microphone array processing for breath monitoring | Multi-microphone recordings | Experimental data | Beamforming, noise filtering |
| Skalicky et al. (2021) | Breath phase-detection for respiratory disease | Lung auscultation sounds | 41 patients | High-pass & low-pass filtering, envelope detection |
| Xu, Yu, and Chen (2020) | Smartphone-based breath monitoring in moving vehicles | Smartphone microphone data | 10 drivers | GAN, Hilbert Transform, ESD analysis |
| Verde et al. (2023) | AI-based voice classification for COVID-19 detections | Voice samples from Coswara dataset | 166 patients | MFCC, Spectral Roll-off, Random Forest, SVM, CNN |
| Shimizu et al. (2023) | AI-based breath sound anomaly detection | Upper airway breath sounds | 60 patients | Stridor Quantitative Value (STQV) AI processing |
| VR et al. (2023) | Photoacoustic breath analysis for COPD and Asthma | VOCs from exhaled breath samples | 69 subjects, and 3 classes (COPD, Asthma, Normal) | PCA, statistical analysis, ROC curve classification |

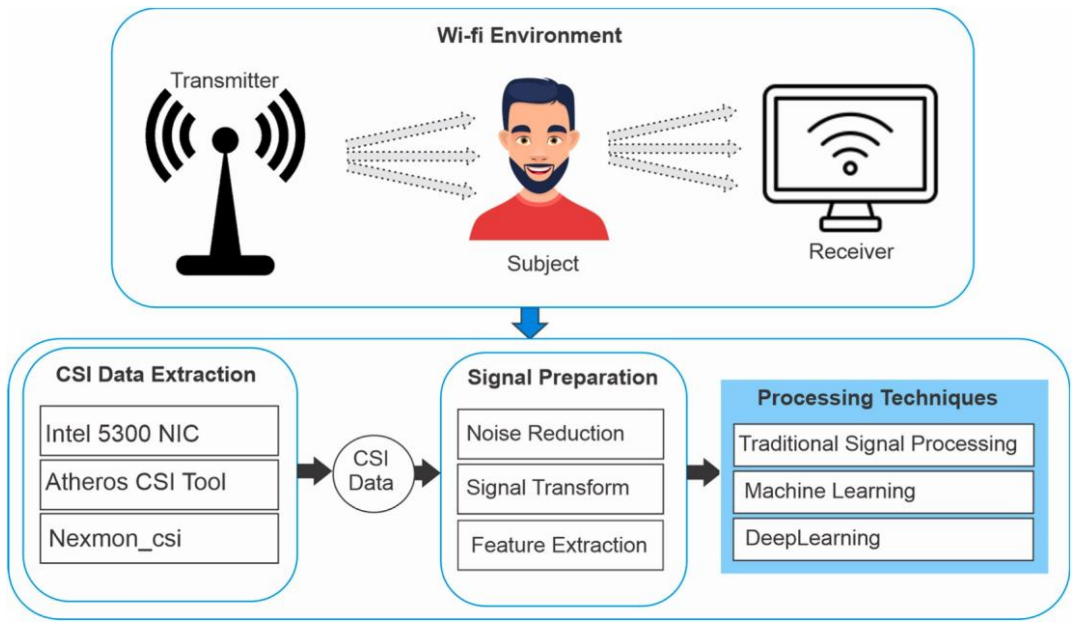

**Fig. 6.** Wi-Fi CSI approach.

filtering for data processing, achieving high tracking accuracy across multiple subjects. Similarly, Wang et al. (Wang, Zhang, et al., 2020) approach involves adaptive subcarrier combination, iterative dynamic programming, and trace concatenation for tracking respiration to count people and recognize individuals using

Wi-Fi CSI. The system's performance is evaluated, and high accuracy is achieved in a lab and a car. Wan et al. (Wan et al., 2023) also focused on tracking respiratory patterns in multi-user scenarios using CSI amplitude sensing with Orthogonal Frequency-Division

**Table 9**
Summary of Wi-Fi CSI-based applications.

| Reference | Approach | Data Type | Data Size | Data Processing Method |
|---|---|---|---|---|
| Guan et al. (2022) | Radiofrequency and phase sensing | Radiofrequency phase signal | Ground truth from 5 subjects | Beamforming AOA_FREQ and DB-SCAN |
| Gao et al. (2020) | CSI sensing of Doppler AOA | CSI data obtained from Wi-Fi with three antennas | Divided into batches | Use DBSCAN for clustering |
| Xu, Guo, and Chen (2020) | CSI sensing for single-user RR detection | CSI signal | N/A | Noise filtering with low-pass filter |
| Brieva et al. (2023) | Single-user RR tracking | Motion-magnified video and MVC | 46800 images of 480x640px | CNN for feature extraction |
| Hu et al. (2022) | Deep Learning RR detection | CSI obtained from COTS devices | 1120 training and 282 test samples | Outlier removal, linear interpolation, and Savitzky-Golay filter |
| Rehman et al. (2021) | CSI sensing for breath abnormality detection | CSI frequency amplitude and phase | Extensive, simulated data for training | Sub-carrier selection, outlier removal, data smoothing |
| Moshiri et al. (2021) | Human activity recognition using deep learning | CSI amplitude | Converted to RGB images | Use pseudocolor to create matrix plots with CNN/BLSTM |
| Dou and Huan (2021) | CSI sensing for RR monitoring | CSI | 1 user scenario | Phase fitting, multipath decomposition |
| Ma et al. (2021) | Wi-Fi CSI deep neural networks reinforcement learning | CSI data | 114 subcarriers, 100 subjects | Captured using Wi-Fi devices; preprocessed with noise filtering and normalization; processed with deep neural networks reinforcement learning. |
| Wang, Yang, and Mao (2020) | Deep learning for activity detection | Human activity data | 10000+ samples | Captured using motion sensors; preprocessed with data augmentation and normalization; processed with CNNs |
| Yin et al. (2021) | Respiration sensing with smartphones | Wi-Fi signals | 100+ respiration cycles | Captured using smartphone Wi-Fi signals; preprocessed with signal filtering and feature extraction; processed with machine learning models. |
| (Gouveia et al. (2024)) | Wi-Fi CSI, Signal Processing | CSI Data | 59 Participants, 17 Activities | Moving Average Filter, Bandpass Filter, PCA, FFT |
| Liu et al. (2020) | Wi-Fi CSI-based analysis | CSI measurements from Wi-Fi devices | 20 participants, 200–300 respirations segments per subject | EMD-based noise removal, periodicity, and sensitivity-based subcarrier selection processed with Waveform morphology analysis, fuzzy wavelet packet transform, DNN-based classification |
| Wang, Zhang, et al. (2020) | Adaptive subcarrier combination and iterative dynamic programming | Wi-Fi CSI | Multiple test scenarios | STFT, subcarrier combination, dynamic programming |
| Srivastava et al. (2021) | Deep Learning | Respiratory Rate Data | Multiple Samples | Deep Learning Techniques |

Multiplexing (OFDM) and FFT. This approach demonstrated robust performance in detecting respiration rates in moving scenarios. In (Ma et al., 2021), the authors leveraged deep neural networks and reinforcement learning for activity recognition using Wi-Fi CSI. This method can be adapted for respiratory pattern detection, offering high accuracy and robustness. The Study (Guo et al., 2022) demonstrated Wi-Fi CSI's capability for user identification based on breath patterns. By employing the Weighted Minimum Distance-Dynamic Time Warping (WMD_DTW) algorithm, their system achieved 97.5 % accuracy, highlighting Wi-Fi CSI's potential in secure user identification and authentication. Liu et al. (Liu et al., 2020) proposed a system leveraging off-the-shelf Wi-Fi devices to capture unique respiratory patterns without requiring active participation from users. They used a Deep Neural Network (DNN) for user verification and spoofing detection, with robust success rates in detecting spoofing attacks. Building on this, the study (Gouveia et al.) explored heart rate monitoring using contactless Wi-Fi CSI data from 59 participants across 17 activities. By carefully tuning signal processing parameters, including moving average filtering, bandpass filtering, PCA, and FFT, they achieved accurate heart rate detection, positioning Wi-Fi CSI as a viable low-cost solution for continuous heart rate monitoring. Lastly, in (Srivastava et al., 2021), the authors proposed a deep learning-based approach for robust respiratory rate sensing using contactless methods. By applying various deep learning techniques, they enhanced the accuracy and reliability of non-contact respiratory rate measurements, making them valuable for continuous health monitoring.

In summary, this section has reviewed the approaches and methods employed in breath analysis, highlighting the limitations of wearable devices and the advantages of contactless approaches. Acoustic signal processing and Wi-Fi CSI methods significantly improve accuracy, user comfort, and data robustness. Among these, Wi-Fi CSI stands out as the most effective approach, providing enhanced accuracy, robustness to noise, and broad area coverage, making it a superior choice for single-user and multi-user respiratory monitoring applications.

## 5. Common architecture

Breath analysis systems generally follow a structured framework involving data acquisition, pre-processing, pattern detection, and classification. Each phase employs specific methods and tools to enhance system performance and accuracy, which is crucial for improving the precision and reliability of breath analysis. This section presents a detailed review of the typical structure associated with breath analysis systems, from data acquisition to pre-processing stages, pattern detection, and classification. Each phase will be elaborated on according to its underlying concept, execution, and importance for improving precision and dependability in breath analysis.

## 5.1. Data capturing and extraction

The first and most critical phase in the breath analysis process is data acquisition, which involves collecting raw signals from wearable and wireless sensors. The quality of the acquired data significantly impacts the accuracy and reliability of the analysis; various approaches to data acquisition include wearable devices, acoustic sensors, and Wi-Fi devices. The next step after the data collection involves extraction and transformation into a structured format; this makes it easily accessible and ready for the subsequent pre-processing operations.

### 5.1.1. Wearable devices

Wearable device methodology embeds various sensors, such as accelerometers, stethoscopes, gyroscopes, and biosensors, to monitor physiological signals related to respiration. These sensors are usually placed on either the chest or abdomen to measure movement caused by respiratory effort. They measure the expansion and contraction in the chest or abdomen during the breathing cycle. Data obtained includes respiratory rate, tidal volume, and breathing patterns.

For example (Islam et al., 2018), used a 4-channel data acquisition system consisting of four dual head stethoscopes (MDF-747) attached with a condenser microphone in the tube of each stethoscope. It was a customized device where a microphone, amplifier, filter, and analog-to-digital converter were used as per the Computerized Respiratory Sound Analysis (CORSA) guidelines. Data from wearable devices is extracted by processing the signals from embedded sensors. Common methods include:

- **External USB sound card (A/D converter):** A/D converter is widely used to convert the output analog signals into digital signals at a sampling frequency of 8 kHz. Digitized data was stored as.wav files in 16 bits, pulse code modulation, and mono audio format in a digital computer through a USB hub for further processing.
- **Time-domain analysis (TDA):** TDA involves analyzing the signal in the time domain to extract features such as peak amplitude, breath duration, and the time interval between breathing. It focuses on the signal's time variations.
- **Frequency-Domain Analysis (FDA):** FDA is a method to decompose a time-domain signal into its frequency-domain counterparts using Fourier transforms to search for periodic patterns linked with breathing.
- Multi-sensor fusion (MSF) provides higher levels of precision by offering a more holistic perspective on respiratory activities through integrating information obtained from various sensors, namely, accelerometers and gyroscopes.

For example, Pharm et al. (Pham et al., 2022) utilized multi-sensor Fusion by integrating Audio, Accelerometer, and Gyroscope information. They use the Audacity tool for Annotation and process the data using CNN-LSTM and TCN, achieving high performance in various scenarios. Park et al. (Park et al., 2023) used tissue hydrodynamic response data with the MSF and FDA to accurately classify breath patterns. Islam et al. (Islam et al., 2018) utilized multi-channel lung sound signals passed through an analog signal conditioning unit that amplified the signals and reduced the environmental noise by filtering using the FDA technique. The output is then converted into a digital signal using the External USB sound card (A/D converter). Tran et al. (Tran & Tsai, 2019) compare traditional and stethoscope-acquired data and apply the TDA and FDA with Noise reduction and signal enhancement. Tran et al. (Tran et al., 2023) capture bronchial breath sounds at optimal neck positions and use the FDA to find the periodic patterns linked with breathing.

### 5.1.2. Acoustic sensors

Acoustic sensors capture the sounds of breathing with microphones kept near the subject. They are noninvasive and can monitor breath patterns continuously (Wang et al., 2023). These sensors detect sound waves caused by the passage of air through the respiratory tract upon inhalation and exhalation. Further analyses of captured sounds provide the respiratory parameters. Common acoustic sensors include standard microphones widely used for recording breath data in various devices. Stethoscope microphones, typically found in digital stethoscopes, are specially designed to capture lung and heart sounds, and thus, they work efficiently in detecting breathing patterns. Piezoelectric sensors also measure mechanical vibrations caused by the breath sounds. Respiratory data can be recorded using smart speakers like Amazon Echo or Google Home, which have built-in microphones. Advanced configurations include microphone arrays, which are multiple microphones in strategic positions that enhance accuracy by reducing ambient noise and increasing the clarity of breathing acoustics. Ultrasound microphones, capable of sensing high-frequency sound waves, detect subtle changes in breath sounds and are used in clinical evaluations. These acoustic sensors enable continuous, non-invasive monitoring of respiratory functions from various applications and operational settings (Kong et al., 2024; Mallegni et al., 2022; Sfayyih et al., 2023; Wang et al., 2018a, 2018b). Once the data is captured by the acoustic sensors, different data extraction methods are employed to convert the analog sound signals to a digital form for further analysis:

- **Analog-to-Digital Conversion (ADC):** First, the continuous auditory signal is converted into a digital form by sampling at regular intervals. It converts the sound waves into a series of numbers, each representing information about the signal.
- **Sound Wave Analysis:** Analyze the amplitude and frequency of sound waves to identify breathing events. It captures variations in sound intensity and pitch caused by breathing.
- **Spectrogram Analysis:** Visualizes the spectrum of frequencies in the audio signal over time, aiding pattern recognition. It creates a visual representation of the signal's frequency content.

For example, Gong et al. (Gong et al., 2022) developed BreathMentor, which uses acoustic sensors to capture diaphragmatic breathing sounds for classification. Similarly, Wang et al. (Wang et al., 2023) use acoustic signals from smart speakers and mic arrays for multi-user heartbeat monitoring.

### 5.1.3. Wi-Fi CSI

Wi-Fi CSI (Channel State Information) captures fine-grained channel information, which can infer breathing patterns by analyzing the impact of respiration on wireless signals. Wi-Fi CSI measures the amplitude and phase of Wi-Fi signals between a transmitter and receiver. Respiration causes subtle changes in these signals, which can be detected and analyzed to determine respiratory parameters.

Various tools capture and extract CSI data, providing the necessary raw information for analysis. One of the pioneering tools is the Intel 5300 NIC (Halperin et al., 2011), which enables the extraction of the CSI data with the support for 30 subcarriers per pair of antennas on a bandwidth of 20 MHz. The Atheros CSI Tool (Xie et al., 2015) does better and supports 114 subcarriers in the case of a pair of antennas on a 40 MHz bandwidth; hence, it can serve more advanced scenarios.

Another well-known tool is the Nexmon CSI Tool (Schulz et al., 2017), which enables extraction on mobile phones and embedded devices, including the Raspberry Pi; it enables multiuser detection and using the CSI data of up to 256 subcarriers for antennas running on an 80 MHz bandwidth. The data from CSI is on different platforms and can be highly adaptable to various experimental setups.

The ESP32 CSI Tool (Hernandez & Bulut, 2020b) represents a lighter alternative, making it possible to extract the raw CSI data directly from the ESP32 microcontroller. Commercial Off-The-Shelf Devices: Standard Wi-Fi routers and smartphones can capture CSI data using specific firmware modifications, enabling real-time breath monitoring in nonspecialized environments. Some Software-Defined Radio platforms, such as the Universal Software Radio Peripheral and the Wireless Open Access Research Platform, can be used for advanced research and support CSI measurements in different frequency bands, including 2.4 GHz, 5 GHz, and even 60 GHz (Jeong et al., 2020).

These tools are adaptable to various research environments, allowing CSI data to be captured and extracted from different platforms for further analysis.

For example, Zeng et al. (Zeng et al., 2018) used the Intel 5300 NIC for respiration detection, achieving high tracking accuracy. Wan et al. (Wan et al., 2023) and Guan et al. (Guan et al., 2022) used the Atheros CSI Tool for multi-user respiratory rate detection, demonstrating its effectiveness in complex scenarios. FFT and other signal-processing techniques were applied to extract respiratory features from Wi-Fi CSI data.

In conclusion, the choice of data acquisition method significantly impacts the effectiveness of breath analysis systems. Wearable devices, acoustic sensors, and Wi-Fi CSI each have advantages and limitations, and the selection depends on the specific application requirements and constraints.

### 5.2. Data pre-processing

Data pre-processing is essential for improving raw data quality and preparing it for accurate and reliable analysis. This stage includes filtering, segmentation, normalization, and artifact removal to make the data consistent and suitable for subsequent steps in breath analysis.

### 5.2.1. Filtering and segmentation

Techniques are critical in improving the quality of breath analysis data. Filtering removes noise from the signal, while segmentation divides continuous data into smaller fragments for more focused analysis.

**Relevant Techniques:**

- **Low-Pass Filters:** These are designed to eliminate high-frequency noise from the signal, preserving the low-frequency components related to respiratory patterns. Xu et al. (Xu, Guo, & Chen, 2020) applied low-pass filters to CSI signals, effectively reducing interference and improving clarity for single-user respiratory rate detection.
- **Segmentation Using Sliding Windows:** Breaks down continuous signals into overlapping or non-overlapping segments, facilitating in-depth analysis of each segment. Zeng et al. (Zeng et al., 2018) used segmentation with FFT and Savitzky-Golay filters to process CSI data, improving the accuracy of respiration detection.

We implemented several methods to explore filtering techniques further and visualized their effects on the CSI signal. The filters applied include Butterworth, Savitzky-Golay, Moving Average, and Wavelet Denoising, each offering unique benefits for noise reduction:

**Butterworth Filter:** A bandpass Butterworth filter was used to smooth the amplitude and isolate the relevant frequency range. This filter achieved a smooth passband transition, helping to enhance the breathing signal by removing unnecessary noise.

**Savitzky-Golay Filter:** The Savitzky-Golay filter, a polynomial smoothing filter, maintains the signal's shape and peak features, which is beneficial for retaining fine details within the CSI data.

**Moving Average Filter:** This filter averages the data points within a specified window size, producing a smooth signal highlighting the breathing cycle pattern.

**Wavelet Denoising:** This method utilizes wavelet decomposition to remove noise by thresholding high-frequency components. The wavelet approach is practical for isolating respiratory patterns without distorting the signal.

Each filter preserves critical breathing signal patterns, which is essential for reliable feature extraction and analysis.

To illustrate the impact of various filtering methods on the CSI amplitude signal, Fig. 7 compares the original CSI amplitude with the outputs of four filtering techniques: Butterworth, Savitzky-Golay, Moving Average, and Wavelet Denoising filters. These filters demonstrate their ability to reduce high-frequency noise while preserving the underlying periodic patterns associated with breathing signals. Each method offers unique advantages depending on the desired balance between noise suppression and signal fidelity, as highlighted in the figure.

### 5.2.2. Normalization

Refers to adjusting values from different datasets to a standard scale, mitigating variations caused by signal strength, distance, and environmental noise. This step is essential to ensure consistency in data, allowing models to focus on respiratory patterns without interference.

***Techniques:***

- ***Min-Max Normalization:*** Scales values within a specific range, typically between 0 and 1. This is useful for algorithms sensitive to the scale of input features. The formula for Min-Max normalization is:

$$x' = \left( \frac{x - x_{min}}{x_{max} - x_{min}} \right)$$

where x' is the normalized value, x is the original value, and x_min and x_max are the minimum and maximum values in the dataset. Guan et al. (Guan et al., 2022) utilized Min-Max normalization to standardize CSI data in multi-user respiratory rate detection.

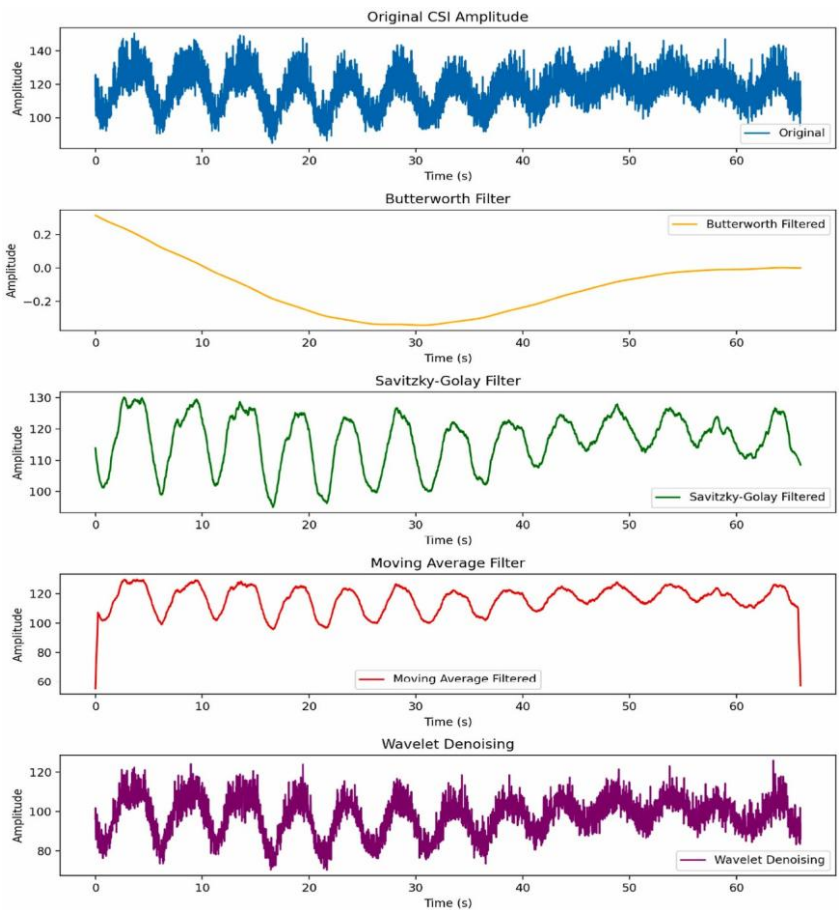

**Fig. 7.** Comparison of original CSI amplitude with outputs from Butterworth, Savitzky-Golay, Moving Average, and Wavelet Denoising filters. Filters effectively reduce high-frequency noise, enhancing the original CSI amplitude.

- **Z-Score Normalization:** Also known as standardization, this technique transforms the data into a mean of zero and a standard deviation of one. It is beneficial for data following a normal distribution. The formula for Z-Score normalization is: $z = (\frac{x - \mu}{\sigma})$ where z is the normalized value, x is the original value, µ is the mean, and σ is the standard deviation of the dataset.

### 5.2.3. Artifact removal

Eliminates non-respiratory signals and interferences that could affect analysis. This ensures cleaner data, especially when subject movements or environmental factors introduce unwanted noise.

> **Techniques:**

- **Time-Based Outlier Detection:** Techniques like the Exponentially Weighted Moving Average (EWMA) smooth signals and remove outliers. Hu et al. (Hu et al., 2022) combined outlier removal with linear interpolation to refine CSI data for reliable respiratory rate detection.
- **Frequency-Based Filtering:** Low-Pass Filtering (LPF) and Wavelet Denoising remove high-frequency noise from the signal, preserving respiratory components. Dou and Huan (Dou & Huan, 2021) used LPF and phase information to enhance CSI data for accurate respiration monitoring.
- **Empirical Mode Decomposition (EMD):** Decomposes the signal into intrinsic mode functions to enable localized noise removal and highlight meaningful patterns. Brieva et al. (Brieva et al., 2023) utilized EMD to extract respiratory features from motion-magnified video data.

Fig. 8 illustrates the impact of artifact removal from CSI amplitude using median filtering, a time-based outlier detection technique. The original CSI amplitude contains irregular spikes caused by noise and system artifacts. The filtering method effectively removes these artifacts, preserving the periodic respiratory pattern essential for accurate breath analysis.

Data pre-processing techniques such as filtering, segmentation, normalization, and artifact removal are essential in preparing raw data for reliable respiratory analysis. Previous studies demonstrate the effectiveness of these techniques in improving the accuracy of breath analysis systems in diverse settings.

### 5.3. Feature extraction

After pre-processing, the next step involves extracting meaningful features from the cleaned data. Feature extraction identifies relevant characteristics from the data for classification and analysis. It extracts significant features from the signal that indicate respiratory patterns, such as peaks, troughs, frequency components, and amplitudes. Feature extraction plays a pivotal role in translating raw signals into meaningful data, which can be used for accurate classification and analysis. This process highlights significant characteristics from the signal, enabling models to detect patterns indicative of respiratory activities.

> **Techniques:**

- **Time-Domain Feature *Extraction*:** Involves extracting features directly from the time-domain signal, such as peak-to-peak interval, breath duration, and amplitude variations.
- **Frequency-Domain Feature Extraction:** Uses spectral analysis to identify frequency components associated with respiratory patterns, such as spectral energy and dominant frequencies.

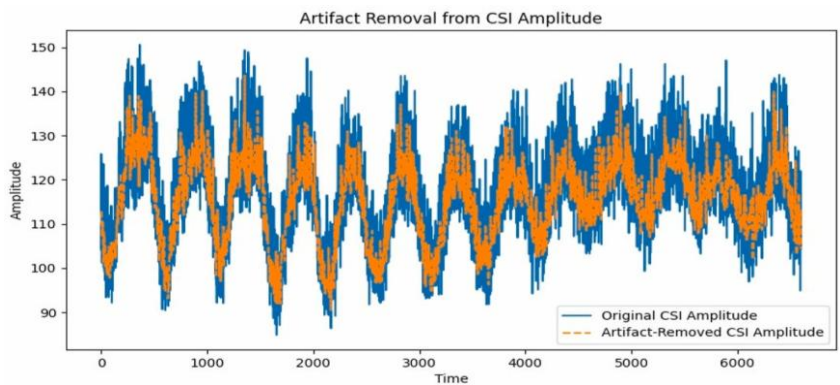

**Fig. 8.** Artifact removal from CSI amplitude using median filtering as a time-based outlier detection technique. The original CSI amplitude (blue) exhibits irregular spikes caused by noise, while the artifact-removed version (orange) retains the primary respiratory pattern, enabling more accurate breath analysis. (For interpretation of the references to color in this figure legend, the reader is referred to the Web version of this article.)

- **Time-Frequency Features *Extraction*:** Combines time and frequency information to view the signal's characteristics using wavelet comprehensively transforms. Time-domain features capture variations in the signal over time, such as the intervals between breaths or changes in breath intensity, which are critical for detecting abnormal breathing patterns.
- ***Amplitude and Phase Analysis:*** Examines changes in the amplitude and phase of Wi-Fi signals to detect breathing. It captures the signal variations caused by the chest and abdomen movement.
- **Time-Frequency Analysis:** Techniques like FFT analyze how frequency components change over time. It provides insights into the dynamic nature of respiratory patterns.
- **Statistical Feature Extraction:** This process extracts statistical features like mean, variance, and correlation from CSI data. These features help quantify the signal characteristics.
- ***ML/DL Algorithms:*** Use algorithms like CNNs to extract features from audio signals automatically. These models learn to identify relevant patterns in the audio data.

For example, Brieva et al. (Brieva et al., 2023) employed time-frequency domain features using the Hermite Transform to extract meaningful patterns from motion-magnified video data, achieving accurate single-user RR tracking. Similarly, Moshiri et al. (Moshiri et al., 2021) utilized frequency-domain and time-domain techniques by converting CSI amplitude data to RGB images, where CNN/BLSTM models facilitated feature extraction for human activity recognition.

Figs. 9 and 10 demonstrate key aspects of feature extraction. Fig. 9 shows the impact of filtering on the CSI signal, improving clarity in the frequency domain, while Fig. 10 illustrates the effectiveness of time-domain features for accurate breath rate estimation.

### 5.3.1. Time-domain feature extraction

The time-domain analysis extracts the respiratory features directly from the signal waveform without converting it to the frequency domain. The features extracted reflect the signal amplitude and duration changes, offering practical information on breathing patterns.

#### Commonly Employed Time-Domain Techniques

- ***Peak-to-peak interval*** extracts features employed in estimating the respiration rate by computing an interval between two successive peaks.
- ***Amplitude Variability*** extracts signal amplitude variations to discriminate between respiration states.
- ***Zero-crossing rate (ZCR)*** is used to get the rate of times a signal crosses zero, which can be used to find changes in respiratory cycles.
- ***Root Mean Square (RMS)*** computes signal energy, indicating breathing intensity.

Time-domain analysis is computationally lightweight and well-suited for real-time applications with low-latency processing. However, it is susceptible to noise and motion artifacts and can result in erroneous respiratory rate estimation (Romano et al., 2023; Wang, Zhang, et al., 2020).

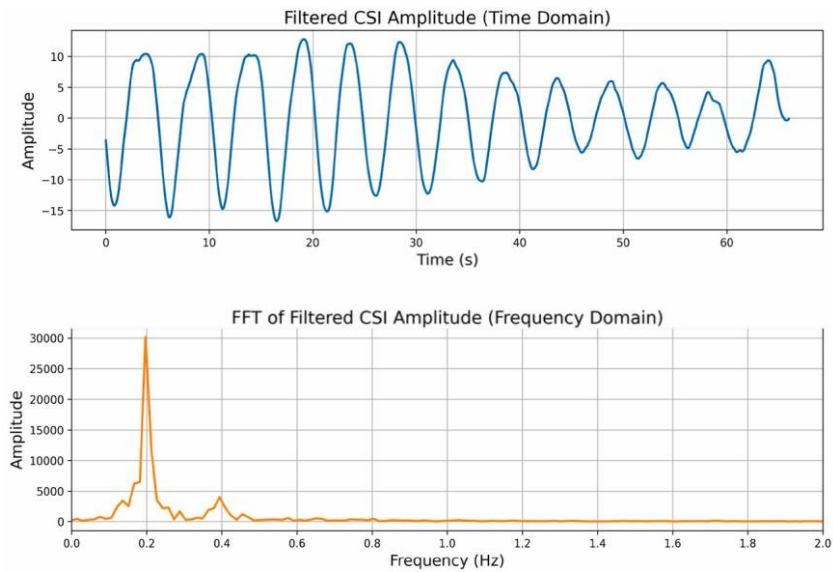

**Fig. 9.** Comparison of filtered CSI amplitudes in time and frequency domains. Filtering enhances feature extraction by reducing noise, allowing the FFT to highlight dominant frequency components related to respiratory patterns.

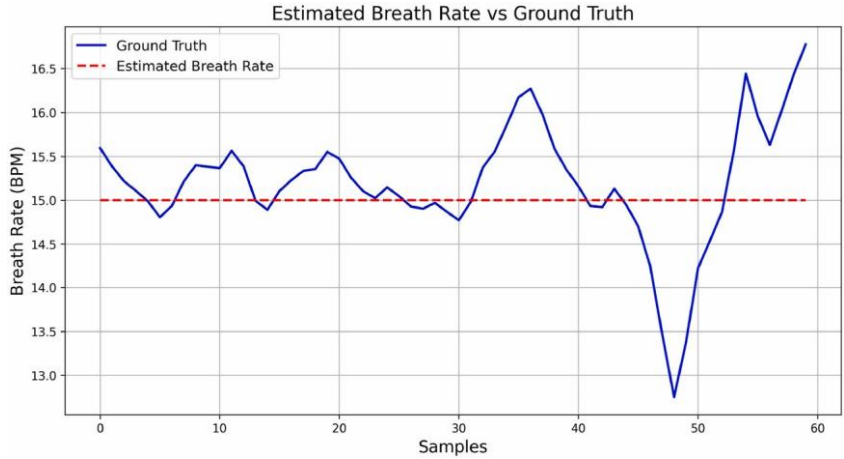

**Fig. 10.** Estimated breath rate versus ground truth. The close alignment of the estimated rate with the ground-truth values illustrates the accuracy of time-domain feature extraction in identifying the user's respiratory patterns.

### 5.3.2. Frequency-domain feature extraction

Frequency-domain analysis transfers time-series data to a frequency spectrum (Harvey et al., 2018), thus accentuating periodic breathing patterns, helping discriminate between different breathing frequencies, and identifying abnormal breathing conditions.

**Typical Techniques:**

- **Fast Fourier Transform (FFT)** converts time-series signals to corresponding frequency components, thus revealing dominant breathing frequencies.
- **Power Spectral Density (PSD)** analyzes the power distribution in the signal over different frequency bands to obtain information about the respiratory rate.
- **Spectral Entropy** measures the frequency distribution complexity to detect abnormal breathing patterns.

Frequency-domain methods are optimum in multi-user respiratory monitoring since the breathing rates of different users can be discriminated against in the spectral domain (Guan et al., 2022; Wan et al., 2023). They also effectively detect abnormalities such as sleep apnea and aperiodic breathing (Indrawati et al., 2022; Molin et al., 2021). Nonetheless, they require extensive preprocessing to remove noise and can miss transient changes in breathing (Harvey et al., 2018).

Fig. 11 illustrates the application of FFT in detecting the dominant respiratory frequency, demonstrating how time-frequency analysis can be employed to monitor and analyze breathing patterns in real time.

### 5.3.3. Time-frequency analysis

Time-frequency analysis methods, i.e., STFT, analyze the time-developing frequency contents of respiratory signals. They look at how the signal's frequency content changes over time and provide information on dynamic respiratory patterns (Guan et al., 2022; Wang, Zhang, et al., 2020). Understanding the time-varying nature of respiratory signals, where breathing patterns may alter over time depending on changes in physiological state, is essential. STFT enables systems to detect dynamic changes in the signal's frequency content, giving more insight into respiratory patterns.

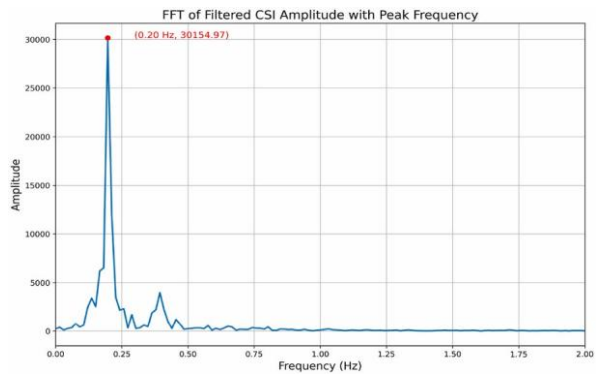

**Fig. 11.** FFT of filtered CSI amplitude with the peak frequency marked, indicating the dominant respiratory frequency. This frequency corresponds to the user's breathing rate, as identified through FFT analysis.

**Commonly Employed Techniques:**

- **Fast Fourier Transform (FFT)** can also convert time-domain signals into frequency-domain representations, identifying dominant frequency components associated with breathing. FFT identifies dominant frequency components, which are crucial in detecting periodic breathing events and making it ideal for continuously monitoring respiratory rhythms (Guan et al., 2022).
- **Hilbert-Huang Transform (HHT)** inherently extracts respiratory oscillations using the Empirical Mode Decomposition (EMD).
- **The short-time Fourier Transform (STFT)** is a time-varying spectral analysis that shows changes in frequency content over time (Wang, Zhang, et al., 2020).
- **Wavelet Transform** facilitates multi-resolution decomposition by separating signals into distinct frequency components at different temporal scales (Brieva et al., 2023). It encompasses methods such as the Continuous Wavelet Transform (CWT), offering high-resolution decomposition suitable for real-time monitoring of respiration, the Discrete Wavelet Transform (DWT), which is particularly useful for feature extraction through dividing signals into a series of hierarchical sub-bands, and the Wavelet Packet Transform (WPT), an extension of DWT providing finer frequency division to identify normal and abnormal respiratory patterns. Such methods augment respiratory monitoring by trading off real-time efficiency against comprehensive spectral analysis.

*5.3.4. ML/DL feature extraction*

Machine learning and deep learning (ML/DL) techniques offer powerful feature extraction through learning complex patterns from raw or minimally processed respiratory data. Unlike traditional methods that utilize hand-crafted features, ML/DL models learn hierarchical and abstract features, which lead to improved classification performance, especially in challenging or noisy conditions.

**Common ML/DL Feature Extraction Techniques:**

- **Convolutional Neural Networks (CNNs):** Capture local spatial features from input representations such as spectrograms, RGB images of CSI amplitude, or time-frequency plots.
- Recurrent Neural Networks (RNNs), LSTM, and BLSTM: Model temporal dependencies in time-series data and are best suited for sequential respiratory signals.
- Autoencoders: Learn compact and noise-resistant feature representations by reconstructing input signals, often used for unsupervised feature learning.
- Hybrid Models: Combine the spatial and temporal feature extraction to handle multimodal or complicated signals.
- Spectrogram-Based DL Features: Convert respiratory signals into time-frequency spectrograms, then apply feature extraction using CNN-based models on image-like input.

For instance, Gong et al. (Gong et al., 2022) used CNNs to extract high-level features from acoustic breathing sounds, achieving high classification accuracy in disease detection tasks. Moshiri et al. (Moshiri et al., 2021) transformed CSI data into RGB images and applied CNN-BLSTM architectures to extract spatio-temporal features for robust activity and respiration classification.

Compared to traditional methods, ML/DL feature extraction techniques improve robustness to noise, eliminate manual engineering, and offer superior generalization across subjects and environments. However, they often require more data and computational resources to train effectively.

*5.3.5. Comparison of feature extraction methods*

Table 10 summarizes the characteristics of time-domain, frequency-domain, and wavelet-based feature extraction techniques to highlight the trade-offs between different methods.

Recent studies, such as (Brieva et al., 2023), have shown that combining wavelet-based features with CNNs improves classification accuracy for breath sound analysis. Similarly, FFT-based spectral features have been used in Wi-Fi CSI-based breath monitoring to distinguish between multiple users effectively [ 6]. Deep learning models trained on spectrograms and RGB representations have proven effective in multi-user Wi-Fi CSI applications (Moshiri et al., 2021).

In conclusion, the choice of feature extraction method should be guided by application requirements. Time domain features are fast and suitable for real-time processing, while frequency and time-frequency methods provide enhanced resolution for complex tasks. ML/DL-based approaches offer the highest accuracy and scalability for intelligent, data-rich respiratory monitoring systems.

*5.4. Classification*

Classification algorithms categorize detected patterns into predefined classes, such as different breathing types or respiratory conditions. This stage is critical for transforming raw data into actionable insights.

*5.4.1. Deep learning-based classifiers*

Deep learning classifiers, such as CNN and LSTM, and their variants, are commonly used for their ability to learn complex patterns from large datasets. They use neural networks to classify patterns based on learned features, providing high accuracy and adaptability to different scenarios.
**Techniques:**

**Table 10**

Comparison of Feature extraction methods for breath analysis.

| Method | Advantages | Limitations | Best Use Case | Impact on Classification Performance |
|---|---|---|---|---|
| **Time-Domain** | It captures direct signal variations, is fast and computationally efficient, and is suitable for real-time applications. | Highly sensitive to noise and motion artifacts, lacks frequency-specific insights. | Real-time single-user respiratory monitoring. | Well-suited for real-time respiratory rate estimation and detecting simple breathing irregularities. |
| **Frequency- Domain** | It highlights periodic breathing patterns, is effective for multi-user detection, and helps detect abnormal breathing conditions. | Requires preprocessing to remove noise and may not capture transient variations. | Multi-user respiratory monitoring, sleep apnea detection. | Improves classification accuracy in multi-user detection, as breathing signals can be separated in the spectral domain. |
| **Time- Frequency Analysis** | It captures both short-term and long-term variations, is robust to noise, and is effective for non-stationary signals. | It is computationally expensive and requires careful parameter selection. | Respiratory disease classification, dynamic breathing pattern analysis. | Wavelet-based methods provide the highest accuracy for respiratory disease classification by capturing dynamic variations in breathing. |
| **ML/DL-Based** | Learns complex features; generalizes across users; minimizes manual effort | Data-hungry; higher computational cost | Multimodal and noisy environments, deep monitoring systems | Achieves state-of-the-art Performance in respiratory classification tasks |

- **Convolutional Neural Networks (CNN):** Specialized in extracting spatial features from data, making them practical for analyzing patterns in images or 2D representations of signals.
- **Long-Short-Term Memory (LSTM) Networks:** These networks are designed to recognize temporal patterns, making them suitable for sequential data such as time-series respiratory signals.

**Hybrid Models:** Combine CNN and LSTM to leverage spatial and temporal feature extraction strengths, enhancing performance in complex scenarios.

For example, McClure et al. (McClure et al., 2020) trained a CNN model to classify human breathing types, achieving variable accuracy across different breathing patterns. Kumar et al. (Chan et al., 2020) compared the performance of various deep learning techniques, including LSTM, BLSTM, C-LSTM, and CNN-LSTM, in tracking respiratory rates and patterns. Moshiri et al. (Moshiri et al., 2021) applied CNN-BLSTM models to CSI data transformed into RGB images, achieving over 95 % activity and respiration classification accuracy. Gong et al. (Gong et al., 2022) used CNNs with transfer learning to detect respiratory anomalies from diaphragmatic sounds in disease monitoring. Pham et al. (Pham et al., 2022) used CNN-LSTM and TCN models on multimodal data for user authentication, achieving up to 95 % accuracy.

### 5.4.2. Machine learning classifiers

Traditional machine learning classifiers, such as SVM and ensemble methods, are also used for their efficiency and effectiveness in specific applications. They utilize traditional machine learning algorithms for classification based on extracted features, offering simplicity and speed in processing. **Techniques:**

- **Support Vector Machines (SVM):** Effective for binary and multi-class classification, SVMs find the optimal hyperplane that separates different classes based on extracted features.
- **Ensemble Methods:** Techniques like Random Forest and Gradient Boosting combine multiple decision trees to improve classification accuracy and robustness.
- **k-Nearest Neighbors (KNN), Naïve Bayes (NB):** Often used in lightweight diagnostic tasks where speed and simplicity are prioritized.

Islam et al. [ 30] employed SVM and ANN classifiers on lung sound data, with ANN showing better performance for asthma classification. Fraiwan et al. (Fraiwan et al., 2021) used ensemble classifiers for lung sound-based disease detection, achieving high diagnostic reliability. Verde et al. (Verde et al., 2023) combined Random Forest and SVM to detect COVID-19 from vocal data with 85–92 % accuracy. In obstructive sleep apnea detection, Molin et al. (Molin et al., 2021) and Indrawati et al. (Indrawati et al., 2022) applied SVM, ANN, and Naïve Bayes to ECG and audio data, showing effective classification with sensitivity exceeding 90 %.

While deep learning models offer higher accuracy and adaptability to complex data, traditional machine learning methods are often more efficient for smaller datasets. This again emphasizes the importance of selecting the proper classification technique according to the exact application.

## 6. Challenges, limitations, and future directions

This section reviews the challenges and limitations of existing systems for breath analysis. It also explores future research directions to address these issues and suggests potential advancements in the field.

### 6.1. Challenges and limitations

#### 6.1.1. Multi-user scenarios

One of the key challenges in breath analysis involves distinguishing between individual respiratory signals in multi-user scenarios. Signal interference and overlap make it challenging to separate users' signals, complicating accurate detection. Guan et al. (Guan et al., 2022) highlighted the importance of beamforming and clustering in improving accuracy in multi-user settings. Beamforming allows signals to be explicitly directed to or from individual users, thus minimizing interference. Clustering algorithms also help distinguish between different signal patterns, enabling a more precise classification of respiratory signals from multiple users.

#### 6.1.2. Impact of subject movements and environmental factors

Subject movements and environmental conditions pose another significant challenge, often introducing noise and artifacts that distort the captured respiratory signals. Movements by the subject can interfere with signal clarity, while environmental noise, such as background sounds, can affect overall data accuracy. Researchers such as Wan et al. (Wan et al., 2023) have developed robust pre-processing techniques and advanced algorithms to mitigate these issues. Using methods such as outlier removal, signal filtering, and advanced ML/DL models, they could distinguish actual respiratory signals from noise, improving the reliability of breath analysis systems.

#### 6.1.3. Environmental factors affecting signal quality

Respiratory monitoring systems, particularly contactless approaches using acoustic sensing and Wi-Fi CSI, are predominantly influenced by environmental conditions such as environmental noise, ambient interference, and signal distortion. These conditions may adversely affect data quality and jeopardize system robustness and accuracy (Guo et al., 2023). Here, we discuss the key environmental challenges and the proposed solutions.

*Effect on Acoustic Sensing:* Acoustic sensing-based breath analysis leans on capturing respiratory sounds via microphones or ultrasonic sensors, and the quality of the signal depends on the environmental conditions, which present various challenges (Romano et al., 2023; Wang et al., 2023):

- *Room Acoustics and Reflections:* Reflections from the walls, ceiling, and objects, as well as echoes, distort the original breath signal, making it challenging to detect breath patterns accurately.
- *Background Noise:* Ambient noise from background sources like a conversation, air conditioning units, or machinery may interfere with detecting breath sounds and cause misclassifications.
- *Microphone Sensitivity and Positioning:* Microphone positioning and sensitivity variations significantly influence the uniformity of the captured breath patterns.

*Mitigation Strategies:*

- *Adaptive Filtering:* Techniques including Wiener or Kalman filtering can suppress background noise while preserving breath- related signals (Khan et al., 2023).
- *Spectral Subtraction:* This technique identifies and subtracts the repetitive background noise in the breath signal, enhancing its quality.
- *Directional Microphones:* Using beamforming techniques or microphone arrays can enhance breath detection by isolating signals from the target subject while minimizing external noise (Romano et al., 2023; Wang et al., 2023).

*Impact on Wi-Fi CSI Sensing:* Wi-Fi CSI-enabled breath monitoring relies on detecting signal variation caused by respiratory movements. However, the following environmental factors can significantly affect the quality of the signal:

- *Signal Interference:* Nearby wireless devices, including Bluetooth, microwaves, and other Wi-Fi networks, can introduce noise that degrades the quality of the captured CSI signals (Wang et al., 2021).
- *Multipath Effects:* Reflections and diffraction of Wi-Fi signals from surfaces such as walls and furniture produce a complex interference pattern, making it challenging to isolate breath-induced variations (Ali, Alloulah, et al., 2021).
- *Human Motion Interference:* Random body movements from other individuals in the environment can interfere with the Wi-Fi signal, compromising the breath detection accuracy (Fan, Pan, et al., 2024).

*Mitigation Strategies:*

- *Multi-Antenna Beamforming:* Directional beamforming techniques can focus the Wi-Fi signal on an individual user, minimizing the effects of environmental interference (Chang et al., 2024).
- *Channel Selection and Adaptive Filtering:* Selecting Wi-Fi channels with less congestion and utilizing filtering techniques (e.g., principal component analysis) can enhance the robustness of CSI-based sensing (Khan et al., 2023).
- *Denoising Algorithms:* Techniques like empirical mode decomposition (EMD) and wavelet denoising can distinguish between respiratory signals and undesired background noise (Alzaabi et al., 2024).

*Considerations for Multi-User Scenarios*: Acoustic and Wi-Fi CSI respiration monitoring systems both face the following challenges in multi-user environments:

- *User Separation:* Signal interference from various users poses challenges in detecting individual breathing patterns with precision (Chang et al., 2024; Wan et al., 2023; Xiong et al., 2020).
- *Dynamic Environment:* Real-world environments have uncontrollable variables that change, such as furniture layout, human motion, and background noise (Albahri et al., 2023; Gui et al., 2022).

*Potential Solutions:*

- *Clustering Techniques,* such as Density-Based Spatial Clustering of Applications with Noise (DBSCAN), may effectively separate the respiratory signals of different users (Chang et al., 2024; Gao et al., 2020; Guan et al., 2022).
- *Hybrid Approaches* combining Wi-Fi CSI and acoustic sensing can increase accuracy by leveraging complementary strengths (Fan, Pan, et al., 2024).
- *Machine Learning-Based Signal Enhancement:* Deep learning models, such as CNN-LSTM hybrids, can be trained to identify respiratory patterns and remove environmental noise (Fan, Yang, et al., 2024).

Addressing these environmental challenges can make breath monitoring systems more robust and reliable, ready for real-world healthcare, fitness, and innovative applications.

*6.1.4. Limited dataset availability*

Another pressing issue in breath analysis is the limited availability of high-quality, annotated datasets. The scarcity of diverse datasets hampers developing and validating more advanced ML/DL models. Most of the available data lacks variety in either demographic or environmental conditions, limiting the generalization of developed models. Data augmentation, which artificially expands the size and diversity of datasets, and synthetic data generation, which simulates data for training, can address these challenges. However, these methods may introduce biases or ethical concerns regarding the authenticity and validity of synthetic data, particularly in healthcare settings where data integrity is paramount. Ensuring that augmented data reflects real-world scenarios remains a significant challenge. Hence, initiatives to encourage open data sharing within the research community would enhance the availability of datasets.

### 6.1.5. Real-time processing and scalability

Real-time processing and scalability are critical to deploying breath analysis systems in practical healthcare applications. However, computational complexity often constrains real-time analysis, as processing breath data requires significant resources. Scalability becomes an issue when systems are expected to process large-scale data or monitor multiple users simultaneously. Solutions to these problems include using edge computing to process data locally, reducing latency, and cloud computing to handle large-scale data processing and storage. Optimization techniques can also balance accuracy with computational efficiency, ensuring systems remain accurate and scalable. Network reliability is necessary for multi-user respiratory monitoring systems since data transmission and sensor network failures can lead to loss of service (Albahri et al., 2023). proposed an intelligent IoT-based mHealth system that enables fault-tolerant healthcare monitoring by direct communication between mHealth systems and distributed hospital networks for continuous patient monitoring.

### 6.1.6. Integration with other health monitoring systems

Integrating a breath analysis system with other health monitoring systems could eventually provide a comprehensive view of a patient's health status. For example, combining breath analysis with heart rate monitoring or oxygen saturation devices in a wearable system could allow for continuous, real-time tracking of key health metrics. However, this integration introduces several interoperability and data integration challenges. The process of consolidating data from different sources, such as wearable or clinical devices, can be complex. Establishing standardized protocols for data exchange and integration, alongside sophisticated techniques for data fusion, will be crucial in facilitating comprehensive health monitoring across diverse systems. *6.2. Future research directions*

### 6.2.1. Open research problems

Despite advancements in breath analysis, significant challenges remain, particularly in scalability and generalizability. As breath analysis systems become more widely adopted, there will be a growing need to scale these systems for large datasets and multi-user environments. Additionally, models must generalize well across diverse populations and settings to ensure reliable and accurate results. For example, breath analysis models developed on a specific demographic may not perform equally well in different geographic regions or among patients with varying health conditions. These challenges reflect broader issues in the healthcare domain, where personalized and adaptable solutions are critical for real-world implementation.

### 6.2.2. Emerging trends in ML/DL approaches

Recent ML/DL trends, including Explainable AI (XAI), federated learning, transfer learning, and hybrid models, have promising directions for further improvement in breath analysis systems. This section addresses these trends with insights regarding their applications and benefits toward breath analysis.

*Explainable AI (XAI).* Breath analysis has been given increasing attention, especially in clinical settings where decisions directly impact patient care. Traditional ML/DL models are usually black boxes that provide predictions without showing the factors driving the projections. XAI addresses this by enhancing model interpretability, allowing healthcare professionals to understand why a model arrives at a particular decision. For example, in a breath analysis model predicting respiratory conditions, XAI can highlight specific features that contributed most to the classification, such as breathing rate patterns or amplitude variations. This interpretability is crucial in clinical decision-making, as practitioners require evidence for the model's outputs before incorporating them into diagnosis or treatment plans. Additionally, XAI fosters trust among patients and healthcare providers, as the underlying logic of predictions is transparent, increasing compliance with model-based recommendations.

*Federated Learning.* (FL) solves the growing need for privacy-preserving breath analysis. Data is often sensitive and decentralized in healthcare institutions, making building comprehensive datasets for model training challenging. FL enables collaborative model training across multiple institutions without requiring data to be centrally stored or shared, thus ensuring data privacy. For instance, hospitals in different locations can participate in training a shared breath analysis model by only sharing model updates rather than raw data. This allows the model to learn diverse respiratory patterns and generalize across demographics, improving accuracy. By addressing privacy and data security concerns, FL facilitates scalable and collaborative development of breath analysis models, opening new opportunities for broader model applicability without compromising patient confidentiality.

*Transfer Learning is instrumental in breath analysis and overcoming* the limitations associated with small or specific datasets. In transfer learning, a model trained for one related task can be fine-tuned for another task with only minimal data. For example, a model initially trained for general respiratory signal analysis can be adapted for more specific tasks, such as sleep apnea detection. This is particularly useful when collecting labeled respiratory data, which is laborious or time-consuming. Transfer learning also reduces the computational resources required for training, so robust breath analysis models can be developed in resource-constrained environments. Similarly, using transfer learning, pre-trained knowledge will help in faster adaptation and better performance, wildly when breath analysis must generalize over different conditions and populations.

*Hybrid Models:* Respiratory signals are very complex; hence, hybrid architecture combining CNNs and RNNs has been promising in handling this complexity. CNNs are more suited for capturing spatial features, such as changes in the amplitude or phase of a signal. At the same time, RNNs are better at processing temporal dependencies, including breathing rhythms over time. For instance, a hybrid model can use a CNN layer to process and filter spatial data from breath signals and an RNN layer to capture temporal patterns that indicate respiratory conditions. This approach is efficient in multi-user scenarios or for detecting irregular breathing patterns where both spatial and temporal information are critical. By leveraging the strengths of both CNNs and RNNs, hybrid models improve the robustness and accuracy of breath analysis systems, making them suitable for complex environments such as home-based health monitoring where diverse breathing patterns may occur.

Each of these trends—XAI, federated learning, transfer learning, and hybrid models—enhances the practical applicability of ML/DL in breath analysis by addressing specific challenges like interpretability, privacy, data scarcity, and complexity. Implementing these trends in breath analysis systems can improve model performance and increase the likelihood of successful deployment in real-world healthcare applications.

### 6.2.3. Recommendations

Future research in breath analysis should focus on interdisciplinary collaboration between experts in healthcare, machine learning, and signal processing. This kind of collaboration will help develop more robust and comprehensive systems. Standardized protocols and datasets should be created to facilitate better training and evaluation of breath analysis models. Privacy and security must also be prioritized, particularly when handling sensitive health data. Longitudinal studies should be conducted to understand how different factors influence the accuracy and reliability of breath analysis systems over time. Lastly, designing user-friendly systems that are minimally intrusive will improve compliance and data quality, enhancing the overall effectiveness of breath analysis technologies.

## 7. Conclusion

With the help of ML/DL techniques, breath analysis has become a robust, non-invasive health monitoring method. This paper surveys various breath analysis applications, including respiratory rate detection, user identification, heartbeat monitoring, and disease detection, providing an overview of the field's recent methods, challenges, and future directions. The survey underlines considerable progress in breath analysis based on technologies like Wi-Fi (CSI), acoustic signal processing, and wearable devices. These methods offer high accuracy in detecting respiratory patterns and show great potential for continuous health monitoring and early disease detection. However, challenges like handling multi-user interference, addressing data scarcity, and improving real-time processing remain key obstacles to broader adoption. To overcome these limitations, future research should prioritize enhancing dataset diversity, refining algorithms for complex scenarios, and integrating breath analysis with other health metrics. Emerging trends such as Explainable AI, federated learning, and hybrid models can offer promising solutions to improve breath analysis systems' scalability, accuracy, and generalization.

In summary, breath analysis can revolutionize health care by enabling continuous, real-time patient monitoring and potentially early diagnostics through interdisciplinary cooperation and innovation to bring the field to life and unlock its full potential for making it a cornerstone of personalized healthcare.

## 8. Funding

This research received no specific grant from any funding agency in the public, commercial, or not-for-profit sectors.

## CRediT authorship contribution statement

**Almustapha A. Wakili:** Conceptualization, Writing – original draft. **Babajide J. Asaju:** Writing – review & editing. **Woosub Jung:** Supervision, Project administration, Validation and recommendations.

## Declaration of competing interest

The authors declare that they have no known competing financial interests or personal relationships that could have appeared to influence the work reported in this paper.

## Data availability

No data was used for the research described in the article.